\RequirePackage{fix-cm}
\documentclass[10pt,a4paper]{article}
\usepackage[T1]{fontenc}
\usepackage[utf8]{inputenc}
\usepackage{amsmath,amssymb,mathtools,bm,amsthm}
\newtheorem{proposition}{Proposition}
\usepackage{graphicx,xcolor,booktabs,tabularx,array,longtable}
\usepackage{enumitem}
\usepackage[numbers,sort&compress]{natbib}
\usepackage{float,flafter,placeins,needspace,etoolbox}
\floatstyle{ruled}
\newfloat{algorithm}{tbp}{loa}
\floatname{algorithm}{Algorithm}
\newcolumntype{Y}{>{\raggedright\arraybackslash}X}
\newcolumntype{L}[1]{>{\raggedright\arraybackslash}p{#1}}
\usepackage{migro_report}

\usepackage{migro_handdrawn}
\usepackage{xurl}
\usepackage[colorlinks=true,linkcolor=blueink,citecolor=blueink,urlcolor=blueink,breaklinks=true]{hyperref}
\hypersetup{pdftitle={MIGRO: A Decoupled and Portable Long-Term Memory System for Language Models},pdfauthor={Mingxi Zou, Yinghui Xu, Jiaxiang Chen, Langzhang Liang, Wei Zhu, Peng Ye, Runkai Zhao, Ruoxi Jiang, Sijin Chen, Song Wang, Zhuo Wang, Lizhen Qu, Zenglin Xu},pdfsubject={Technical Report}}
\title{\raisebox{-4pt}[0pt][0pt]{\includegraphics[height=21pt]{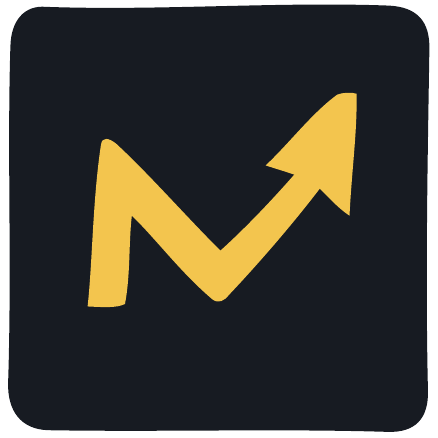}}\hspace{.32em}MIGRO: A Decoupled and Portable\\[2pt]Long-Term Memory System for Language Models}
\newcommand{\affilmark}[1]{\textsuperscript{\textcolor{reportink}{#1}}}
\newcommand{\affillogo}[2][12pt]{\raisebox{-.25\height}{\includegraphics[height=#1]{figures/affiliations/#2.png}}\hspace{.35em}}
\author{%
  \begin{tabular}{@{}c@{}}
    Mingxi Zou\affilmark{1,2},\enspace Yinghui Xu\affilmark{1},\enspace
    Jiaxiang Chen\affilmark{1,8},\enspace Langzhang Liang\affilmark{1},\enspace
    Wei Zhu\affilmark{4},\enspace Peng Ye\affilmark{3}\\[-1pt]
    Runkai Zhao\affilmark{3},\enspace Ruoxi Jiang\affilmark{2},\enspace
    Sijin Chen\affilmark{6},\enspace Song Wang\affilmark{7},\enspace
    Zhuo Wang\affilmark{1},\enspace Lizhen Qu\affilmark{5,\textdagger},\enspace
    Zenglin Xu\affilmark{1,\textdagger}\\[3pt]
    \small \affilmark{1}\,\affillogo{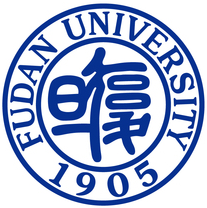}Fudan University\quad
    \affilmark{2}\,\affillogo{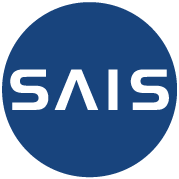}Shanghai Academy of AI for Science (SAIS)\quad
    \affilmark{3}\,\affillogo[10pt]{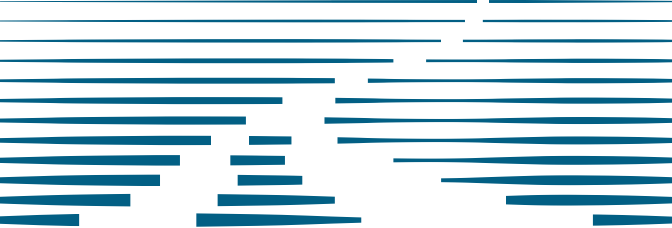}Shanghai AI Laboratory\\[0pt]
    \small \affilmark{4}\,\affillogo{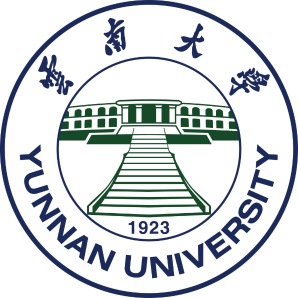}Yunnan University\quad
    \affilmark{5}\hspace{.55em}\raisebox{-.25\height}{\includegraphics[width=40mm]{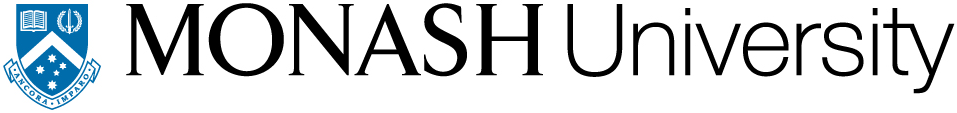}}\quad
    \affilmark{6}\,\affillogo{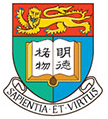}The University of Hong Kong\hspace{.8em}
    \affilmark{7}\,\raisebox{-.2\height}{\includegraphics[height=8pt]{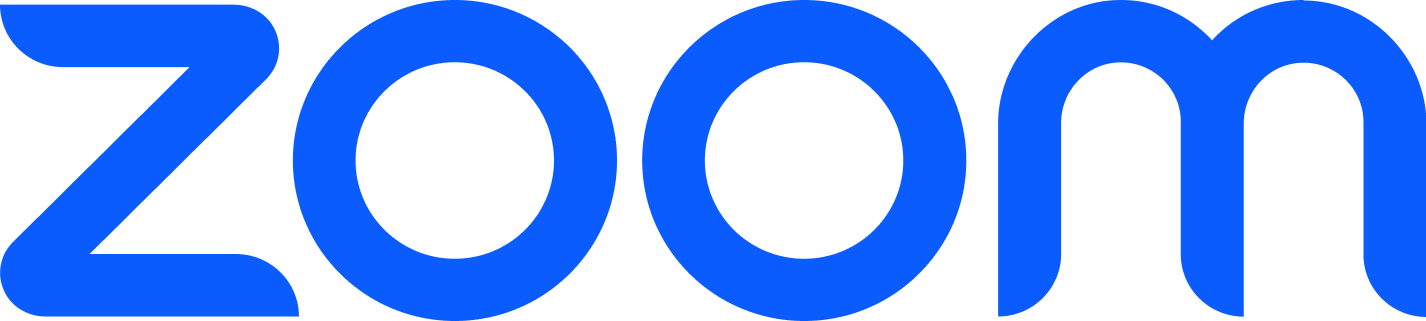}}\\[-1pt]
    \small \affilmark{8}\,Expevo studio\\[-1pt]
    \footnotesize \textcolor{reportink}{\textdagger\enspace Corresponding authors}
    \quad\textcolor{reportgold}{\textbar}\quad Mingxi Zou (\href{mailto:mxzou24@m.fudan.edu.cn}{\nolinkurl{mxzou24@m.fudan.edu.cn}})
  \end{tabular}%
}
\date{}
\begin{document}
\maketitle
\thispagestyle{reportfirst}
\begingroup
\setlength{\intextsep}{5pt}
\begin{figure}[H]
\centering
\includegraphics[width=\linewidth]{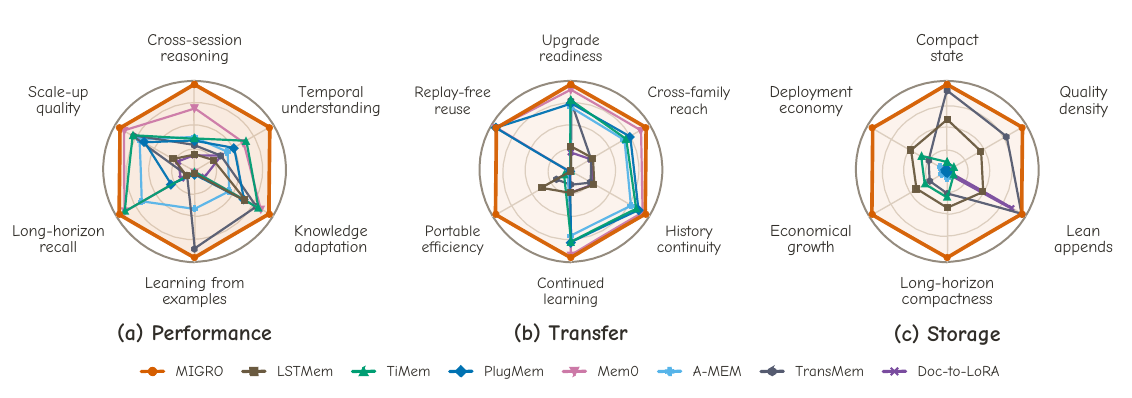}
\captionsetup{skip=4pt}
\caption{MIGRO's capability profiles across performance, transfer, and storage. Farther out indicates stronger measured capability on each axis. Appendix~\ref{sec:joint_details} defines the axes and their scales.}
\label{fig:radar}
\end{figure}
\endgroup
\begin{abstract}
\noindent
Language models evolve, but the observations, revisions, and context accumulated through user interactions should remain usable. Each new model must be able to retrieve past evidence, track revisions, and add new observations. We present \textbf{MIGRO}, a decoupled long-term memory system that treats experience as a persistent state shared across successive language models.

MIGRO follows a simple principle: preserve historical evidence exactly, organize it approximately, and decide what to retrieve only when the question is known. We call this \emph{asymmetric fidelity with deferred selectivity}. A shared semantic index supports retrieval, while exact records retain observations and their historical relationships. Because models differ in vocabulary and tokenization, host-specific interfaces convert selected evidence into inputs the receiving model can use. Reversible encoding reduces archival storage while enabling exact recovery. The shared state and its access interfaces let successive hosts read and extend the accumulated experience.

On matched LoCoMo states, MIGRO scores 7.3\% higher and uses 99.3\% less private memory than the strongest quality baseline. After handoff, continued writing retains 99.9\% of full-history rebuilding quality by extending the inherited state. On million-token BEAM histories, MIGRO uses 57.5\% less private storage on average than the most compact baseline. MIGRO combines high answer quality, continued memory growth across hosts, and low storage cost (Figure~\ref{fig:radar}).
\par\noindent Code is available at \url{https://github.com/zOUMINGXI/MIGRO}.
\end{abstract}

\section{Introduction}
\label{sec:intro}
\textit{Memory should outlive the model.} Long-running user interactions accumulate observations, revisions, and preferences that remain relevant to later questions \citep{locomo,longmemeval}. During those interactions, the language model serving the user may be upgraded or replaced. A receiving model should recover what was said, distinguish earlier versions, and continue accumulating experience. \textbf{The enduring object is the user's history; the reasoning engine around it can change.}

\begin{figure}[!htbp]
\centering
\includegraphics[width=\linewidth]{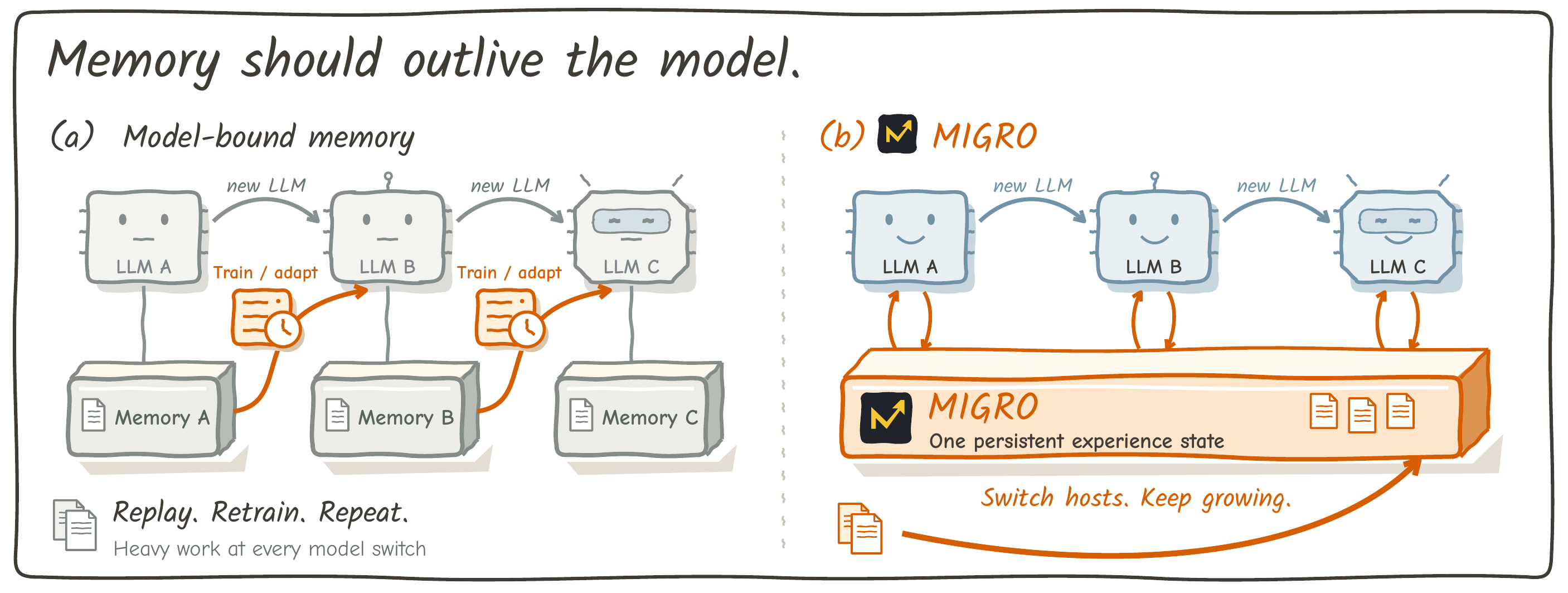}
\captionsetup{skip=4pt}
\caption{An illustrative model-bound approach repeats training or adaptation after host replacement (left). MIGRO lets successive hosts read and extend one persistent experience state (right).}
\label{fig:hero}
\end{figure}

Research on language model memory has advanced along complementary directions. Neural approaches integrate past information through learned representations and model states \citep{wu2022memorizing,metis2026,memt2026,deltamem2026}. Retrieval-augmented and agent memory systems retain explicit evidence beyond the active context \citep{lewis2020rag,memgpt,amem,mem0}, while memory-management frameworks organize evolving records and shared resources \citep{memos2025,evermemos2026}.

External records can already be shared across models, as illustrated by plugin and text-based memory systems \citep{plugmem,mem0}. An earlier revision may be lost when memory is summarized, while surviving evidence may still be difficult for a new host to retrieve or interpret \citep{modelupgrade2026}. Preserving history and making it usable are therefore distinct requirements. Successive revisions may share a semantic neighborhood while answering different questions about the past. A durable memory must retain those distinctions before knowing which ones will matter, including their source and historical relationships \citep{cheney2009provenance,snodgrass1999temporal}.

This motivates two complementary principles. \emph{Asymmetric fidelity} preserves historical evidence exactly while allowing approximate semantic addresses that index observations for retrieval \citep{jegou2011pq,karpukhin2020dpr}. \emph{Deferred selectivity} postpones evidence selection until a question is known. At read time, an evidence packet provides a temporary view of the observations or exact spans selected for that question. Future readers can revisit the retained history with new questions instead of relying on an interpretation fixed at writing time.

We introduce MIGRO, a decoupled long-term memory system that realizes this separation through one persistent experience state and host-specific interfaces. Writing retains canonical observations, their provenance, historical order, and available reference links; reading selects relevant evidence and adapts it to the receiving model's vocabulary and tokenization \citep{kudo2018sentencepiece}. Reversible predictive coding stores this state in an archive, its physical persistent form, from which the saved records can be recovered \citep{witten1987arithmetic,compression}. Changing the host replaces the access interface; subsequent writes extend the inherited state.

The experiments examine three consequences of this design across conversational, operational, and long-horizon memory tasks \citep{locomo,longmemeval,memops,memoryagentbench,beam}. Access ablations reveal how semantic similarity alone can miss complementary evidence. Handoff and continued-writing studies show that inherited history can remain useful and continue growing under new hosts. Matched-state and long-history measurements show that exact preservation can coexist with compact states and low marginal storage cost.

Our contributions are organized around three complementary requirements of long-term memory:
\begin{enumerate}[leftmargin=*,itemsep=4pt,topsep=4pt]
\item \textbf{High-quality memory through fidelity-aware access.} We introduce \emph{asymmetric fidelity with deferred selectivity}, separating exact historical evidence from approximate semantic organization and query-conditioned access. Shared semantic anchors, observation-specific innovations, and contextual evidence retrieval realize this principle, supported by a formal characterization of information preservation and semantic ranking stability.
\item \textbf{Portable memory with continuous experience accumulation.} We develop a decoupled experience state that successive language models can access and extend through host-specific interfaces. Lexical/subword-aware transport enables reuse across heterogeneous models, while continued writing after handoff preserves the continuity of accumulated experience without rebuilding its history through the receiving backbone.
\item \textbf{Storage-efficient memory with exact recoverability.} We combine compact semantic addressing with reversible causal coding to preserve historical evidence at low persistent cost. Long-history measurements, matched quality--storage comparisons, and shared-resource accounting demonstrate how exact preservation can coexist with economical memory growth and scalable reuse.
\end{enumerate}

\Needspace{6\baselineskip}
\section[System Overview]{\migroicon[12pt]\hspace{.38em}System Overview}
\label{sec:overview}
MIGRO represents experience, the accumulated history of observations, as one persistent state. Each observation is a recorded input that contributes to this history. In Figure~\ref{fig:complete}, follow writing from incoming observations into the shared state, then reading from that state to the current host. Persistence makes the state recoverable, while handoff changes the host without replacing the history.

\begingroup
\setlength{\intextsep}{9pt}
\begin{figure}[H]
\centering
\includegraphics[width=\linewidth]{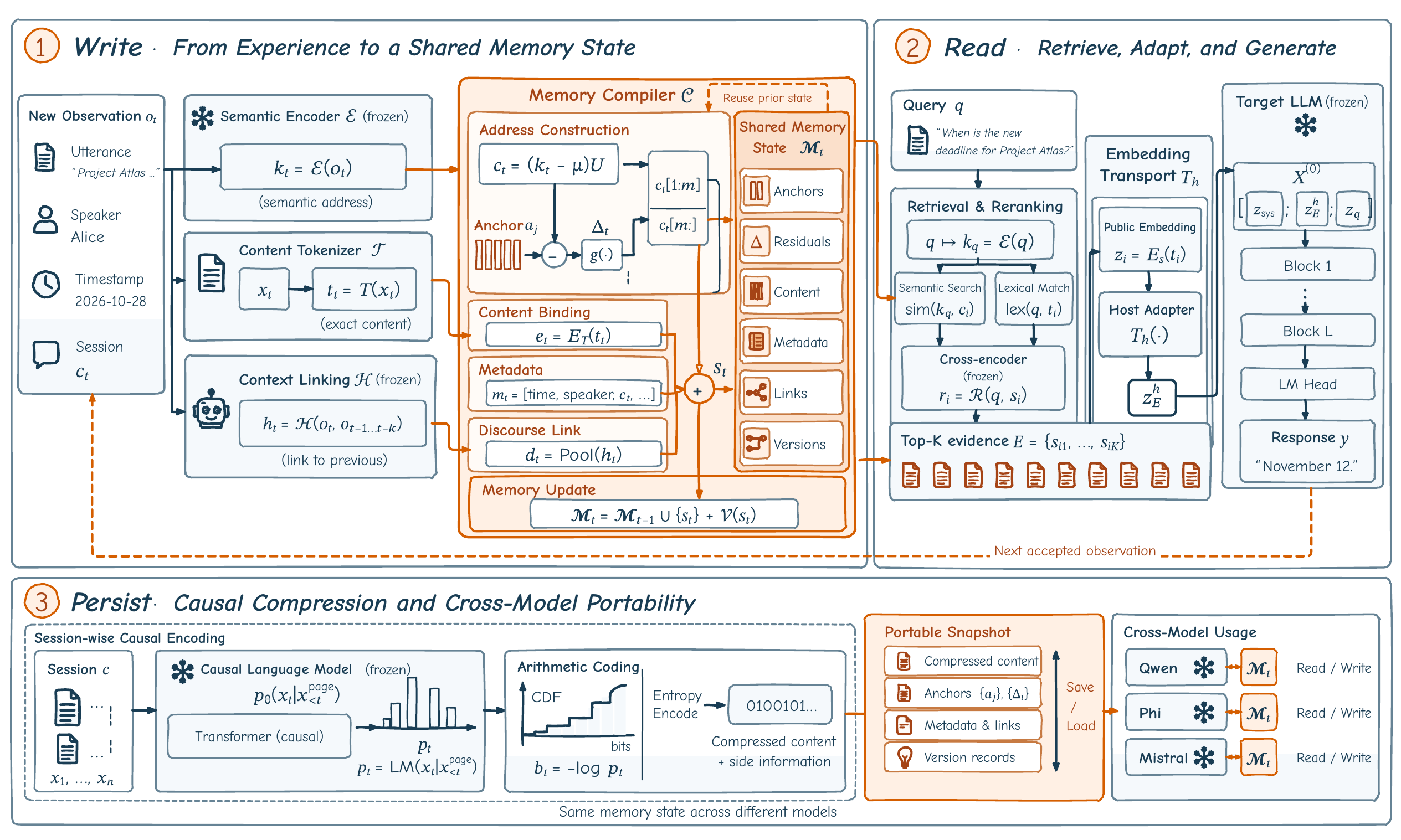}
\captionsetup{skip=5pt}
\caption{One experience state across writing, reading, persistence, and model handoff. Writing adds evidence and addresses; reading constructs a bounded evidence packet; persistence preserves the saved state for recovery. A receiving host reads and extends the inherited history. The diagram uses row vectors; equations use column vectors.}
\label{fig:complete}
\end{figure}
\endgroup

\subsection{A Fact Revision Across Model Replacements}
A user states that a project's budget is 17 in version 1, then revises it to 18 and 19 under successive hosts. \emph{What is the latest budget?} and \emph{What was the budget in version 1?} require different answers, 19 and 17, from the same evolving history. When the original host returns, both questions should remain answerable. Table~\ref{tab:example} in Appendix~\ref{sec:trace_details} expands this illustrative trace.

The example separates three ideas. Semantic similarity does not make observations identical: all three statements concern the same budget but record different values. The latest answer does not replace the full history: retaining only 19 erases the evidence for the first version. Model replacement does not change historical identity: a receiving host inherits earlier records and adds its own observations to that history.

Semantic addresses organize these related records into a common neighborhood. Independent source bindings retain their content, provenance distinguishes their origins and historical positions, and query-time selection determines which version supports the answer. Local discourse links add contextual access hints within a bounded same-session window; cross-session revisions can remain unlinked while their evidence is still preserved.

\subsection{One State, Four Operations}
Writing adds an observation, its address, and its historical relationships to the durable state. Reading leaves that state intact and constructs a temporary, attributed evidence packet for the current question and host. Persistence encodes the saved state so that its records and access structure can be recovered. Handoff attaches a new host interface to this state, allowing the receiver to read inherited observations and continue writing. Algorithm~\ref{alg:lifecycle} in Appendix~\ref{sec:lifecycle_algorithm} connects these operations.

What must the shared state preserve so that these operations remain valid across future queries and hosts?

\section{Problem Formulation and Design Principles}
\label{sec:problem}
Long-term memory must preserve an evolving history while allowing changing models to use it. The budget example exposes both requirements: the state must retain distinct revisions, and a reader must select the revision relevant to its question. We formalize this separation before deriving the design principles.

\subsection{Experience as a Persistent State}
What must persist when the host changes? MIGRO separates three roles: exact evidence retains source-token bindings, provenance, and history; access structure organizes observations through semantic anchors and address innovations; contextual relationships retain references between records.

An accepted observation is $o_i=(x_i,m_i)$, where $x_i$ is its content and $m_i$ contains available provenance, including speaker, role, time, session, and source identity. Observations arrive chronologically and may be written by different hosts $h_i$, each with frozen parameters $\theta_{h_i}$. Shared canonicalization $\sigma$ normalizes each accepted observation, and a source tokenizer $\tau_s$ yields $v_i=\tau_s(\sigma(o_i))$. We call this saved content representation \emph{canonical evidence}. Throughout, \emph{exact recovery} means recovery of these saved source-token IDs and their associated saved state; it does not mean unconditional reconstruction of input forms discarded before canonicalization.

After $n$ observations, the persistent state is
\begin{equation}
\mathcal M_n=(\mathcal A_n,\mathcal D_n,\mathcal X_n,\mathcal P_n,\mathcal G_n,\mathcal V_n).
\label{eq:state}
\end{equation}
Here $(\mathcal X_n,\mathcal P_n,\mathcal V_n)$ hold source bindings, provenance, and retained history; $(\mathcal A_n,\mathcal D_n)$ hold anchors and address innovations; and $\mathcal G_n$ holds reference relations. Shared dependencies $\mathcal B$ specify how to represent source tokens, reconstruct addresses, and decode archives across hosts. The observations remain user-specific; these dependencies are reusable resources.

\subsection{Memory Lifecycle and Model Handoff}
Writing changes what future models can remember. Reading changes what the current model sees. A write extends persistent history; a read constructs a temporary evidence packet for its query:
\begin{equation}
\begin{aligned}
\mathcal M_{i}&=\operatorname{Write}_{h_i}(\mathcal M_{i-1},o_i;\mathcal B),\\
(y,\mathcal E_q)&=\operatorname{Read}_{h}(q,\mathcal M_n;\Pi_h,\mathcal B).
\end{aligned}
\label{eq:operations}
\end{equation}
The access profile $\Pi_h$ specifies selection, the source-token budget, target context allowance, prompt and generation policies, and optional calibration. A record omitted from $\mathcal E_q$ remains in $\mathcal M_n$ for later readers and questions.

Persistence encodes this state, and handoff changes its consumer:
\begin{equation}
\begin{aligned}
S&=\operatorname{Archive}(\mathcal M_n;\mathcal B),\qquad
\operatorname{Restore}(S;\mathcal B)=\mathcal M_n,\\
(\mathcal M_n,h)&\xrightarrow{\operatorname{Handoff}(h')}
(\operatorname{Restore}(S;\mathcal B),h').
\end{aligned}
\label{eq:handoff}
\end{equation}
Restoration recovers saved source IDs, quantized coordinates, provenance, reference relations, and history. The receiver attaches its access interface and appends subsequent observations through the same schema. Appendix~\ref{sec:operational_contracts} specifies these contracts.

Handoff avoids reconstructing full history through the receiving backbone \citep{modelupgrade2026}. Restoration, index preparation, and interface initialization remain separate costs; Appendix~\ref{sec:runtime_complexity} describes them. The new host processes selected evidence during reading and a bounded historical window during writing, while the accumulated state remains continuous.

\subsection{Deferred Selectivity: Why Preservation Precedes Access}
\label{sec:theory}
\label{sec:deferred}
Long-term memory faces an asymmetry in when information becomes relevant. At write time, the questions that will depend on an observation are unknown. At read time, the question is available and evidence can be selected accordingly. Persistence must retain distinctions that future questions may require; reading must expose the relevant distinctions within the current model's access budget.

Let $R_n=((v_i,\bar m_i,i))_{i=1}^{n}$ denote the accepted canonical history, including source-token sequences, saved historical metadata, and observation order. For a family of future questions $\mathcal Q$, let $\operatorname{Ans}(q,R)$ be the answer determined by history $R$. Two histories are equivalent when no question in that family distinguishes them:
\begin{equation}
R\sim_{\mathcal Q}R'
\quad\Longleftrightarrow\quad
\operatorname{Ans}(q,R)=\operatorname{Ans}(q,R')
\quad\text{for all }q\in\mathcal Q.
\label{eq:queryequivalence}
\end{equation}
This relation identifies what a memory must preserve for the given family. The encoding--query separation connects information-theoretic compression to communication complexity \citep{cover2006information,tishby1999bottleneck,kushilevitz1997communication}.

\begin{proposition}[Irreversibility of query-distinguishable information loss]
\label{prop:irreversibility}
Let $F$ be a query-independent memory encoding. If $F(R_1)=F(R_2)$ but $R_1\not\sim_{\mathcal Q}R_2$, no readout using only $q$ and $F(R)$ can guarantee the correct answer to every question in $\mathcal Q$ for both histories.
\end{proposition}

In the running example, irreversibly overwriting the version-1 budget 17 with 19 removes the evidence needed to answer the version-1 question. A stronger future model cannot recover that value from the current state alone. By contrast, a record omitted from today's evidence packet remains available to another query. Deferred selectivity preserves the possibility of reinterpreting history before choosing what to expose on each read. The binary-fact example in Appendix~\ref{sec:binary_example} develops this distinction.

This separation becomes more valuable as future readers acquire new capabilities and information needs. For nested question families,
\begin{equation}
\mathcal Q_1\subseteq\mathcal Q_2,\quad R\sim_{\mathcal Q_2}R'
\quad\Longrightarrow\quad R\sim_{\mathcal Q_1}R'.
\label{eq:familyrefinement}
\end{equation}
The broader family refines the distinctions that memory must retain. An encoding sufficient for an earlier task can merge histories that a later task separates. Canonical history preserves the opportunity to revisit those observations under new questions and stronger models.

\subsection{Asymmetric Fidelity: Exact Evidence, Approximate Addresses}
\label{sec:fidelity}
A book's contents must be preserved, but its index need not encode enough information to reconstruct every page. The same distinction applies to memory: source bindings determine which historical record is preserved, while semantic addresses help locate it. Compact access geometry can therefore tolerate approximation without merging record identities.

MIGRO assigns different fidelity requirements to these roles. The components $(\mathcal X_n,\mathcal P_n,\mathcal V_n)$ retain canonical evidence and its historical context. The components $(\mathcal A_n,\mathcal D_n)$ organize semantic access through quantized shared representations, while $\mathcal G_n$ retains explicit reference relations. Let $\widehat R_n$ be the canonical history recovered from an exactly restored saved state, $k_i$ the original normalized address, and $\widehat k_i$ its reconstructed, normalized address. Then
\begin{equation}
D_{\rm evidence}:=\mathbf 1[\widehat R_n\ne R_n]=0,
\qquad
D_{\rm address}:=\max_{1\le i\le n}\|k_i-\widehat k_i\|_2.
\label{eq:asymmetricfidelity}
\end{equation}
The persistent evidence is recovered exactly while the access geometry admits approximation. Related observations can share a semantic neighborhood yet retain independent values, sources, and historical positions. Quantized vector search provides the technical background for compact semantic coordinates \citep{jegou2011pq,babenko2014aq}.

Quantization may move addresses slightly without changing which records are retrieved when their relevance scores are sufficiently separated. To characterize this condition for dense semantic scoring \citep{karpukhin2020dpr}, consider a fixed candidate set $\mathcal C$, with $s_i(q)=\langle k_q,k_i\rangle$ and $\widehat s_i(q)=\langle k_q,\widehat k_i\rangle$, where $\|k_q\|_2=1$.

\Needspace{10\baselineskip}
\begin{proposition}[Semantic Top-$K$ stability]
\label{prop:topkstability}
Suppose $|\mathcal C|>K$ and $\max_{i\in\mathcal C}\|k_i-\widehat k_i\|_2\le\epsilon$. Write $s_{(K)}(q)$ and $s_{(K+1)}(q)$ for the boundary scores under the original addresses. If
\begin{equation}
\gamma_q:=s_{(K)}(q)-s_{(K+1)}(q)>2\epsilon,
\label{eq:margincondition}
\end{equation}
the semantic Top-$K$ set is unchanged by address reconstruction.
\end{proposition}

Each similarity score changes by at most $\epsilon$. When the original boundary margin exceeds the combined perturbation of competing candidates, their semantic Top-$K$ membership is preserved. Appendix~\ref{sec:proposition_proofs} proves both propositions.

Here $\epsilon$ bounds reconstructed, normalized address error within $\mathcal C$; the reuse criterion in Equation~\ref{eq:address} controls anchor selection separately. This result characterizes the geometric condition under which approximate semantic addressing preserves retrieval membership, while exact source bindings maintain historical identity independently of address approximation.

\subsection{Portable Access and Joint Evaluation}
\label{sec:portable_objective}
All hosts access the same saved observation fields and source vocabulary, but their input representations differ. A host-specific interface selects an evidence packet and maps its source tokens into the receiver's vocabulary and embedding space. Optional calibration belongs to this interface. Host replacement changes how the records enter the model while preserving their historical identities.

For a receiving host $h$, selection and interpretation factorize as
\begin{equation}
\begin{aligned}
\mathcal E_q^{(h)}&=\operatorname{Select}_{\Pi_h}(q,\mathcal M_n),\\
Z_q^{(h)}&=E_s(v_{\mathcal E_q^{(h)}}),\\
y_h&=\operatorname{Gen}_{\theta_h}\!\left(\mathcal I_h(q,Z_q^{(h)};\eta_h)\right).
\end{aligned}
\label{eq:theoryhostfactorization}
\end{equation}
The shared embedding table $E_s$ represents selected source IDs, $\mathcal I_h$ constructs the receiving input, and $\eta_h$ denotes optional host-side calibration. Tokenizer compatibility and target-reader alignment affect how effectively the receiver uses inherited evidence \citep{kudo2018sentencepiece,frozenmemory2026}. The same state remains available for subsequent reads and writes.

This separation calls for joint evaluation along several dimensions:
\begin{equation}
\bigl(Q_h,\ \Delta_{w\to h},\ B_{\rm user},\ B_{\rm shared},\ C_{\rm serve},\ C_{\rm restore}\bigr).
\label{eq:objective}
\end{equation}
Here $Q_h$ measures task quality under a declared access profile, and $\Delta_{w\to h}$ is additional target-relative loss after write trajectory $w$. The remaining terms measure private persistent bytes, shared assets, serving resources, and restoration cost. These are evaluation dimensions rather than a scalar objective: quality tests effective access, transfer tests continuity across consumers, and accounting measures the cost of retaining and using that experience.

The remaining design question is how to give changing hosts useful access to this preserved history at low persistent cost.

\section{MIGRO Architecture and Lifecycle}
\label{sec:design}
The architecture must locate preserved observations, deliver them within a host's input budget, and store the growing history economically. MIGRO uses semantic addresses for lookup, evidence packets for query-specific access, and archives for persistence (Figure~\ref{fig:complete}).

\subsection{Shared Addresses and Exact Evidence}
Related observations should be easy to retrieve together without losing their differences. In the budget example, the three revisions belong to one semantic neighborhood, but each must retain its own value and historical position \citep{cheney2009provenance,snodgrass1999temporal}. A common center alone would flatten their individual placement. An anchor captures what related observations share; an innovation captures where each observation lies within that shared semantic structure. Neither replaces the underlying evidence.

Given a normalized public address $k_i$, shared center $\mu$, coordinate basis $U$, and innovation subspace $W_\delta$, the writer computes
\begin{equation}
u_i=U^\top(k_i-\mu),\quad
e_{ij}=\|(I-W_\delta W_\delta^\top)(u_i-A_j)\|_2.
\label{eq:address}
\end{equation}
The residual $e_{ij}$ measures the difference outside the innovation subspace. When the best anchor satisfies the fixed reuse threshold, the writer saves a quantized local placement and reconstructs the address as
\begin{equation}
\widehat\delta_i=Q_\delta(W_\delta^\top(u_i-A_{a_i})),\qquad
\widehat k_i=\operatorname{norm}\bigl(U(A_{a_i}+W_\delta\widehat\delta_i)+\mu\bigr).
\label{eq:innovation}
\end{equation}
Otherwise, it creates a new quantized root with no additional innovation. Anchor references, coordinates, and scales reconstruct approximate access geometry, while independent source bindings preserve evidence \citep{jegou2011pq,babenko2014aq}. The quantized innovation retains individual placement without promising recovery of the original continuous address.

Writing also adds contextual relationships. The current frozen host processes the incoming observation with a bounded window of earlier same-session records and infers a predecessor from their intermediate representations. All spans are computed within the current host, including records written by another model. Only the observation reference persists; hidden representations remain temporary. Appendix~\ref{sec:reference_inference} gives the rule.

An accepted write appends its address, source binding, provenance, and available references, leaving earlier revisions independently identifiable. This design raises two empirical questions: do innovations improve access quality, and how much address storage do they save? Sections~\ref{sec:quality} and~\ref{sec:storage} address them through the quality and address-table comparisons in Figure~\ref{fig:ablation}d,f.

\subsection{Budgeted Retrieval and Host Adaptation}
Reading has two tasks: choose the evidence, then deliver it to the receiving model. Retrieval can fail by missing a relevant record or its supporting context. Delivery can fail even with the right records because models use different vocabularies, token boundaries, and embedding spaces.

Retrieval constructs a bounded packet from the accumulated state. Reconstructed-address similarity and lexical matching generate candidates, followed by frozen cross-encoder reranking and diversity-aware selection of $K$ primary relevance seeds \citep{karpukhin2020dpr,mmr}. Chronological reserves keep older observations eligible, while local and reference-linked support can supplement the seeds. The final packet can therefore contain more than $K$ records. Appendix~\ref{sec:complementary_selection} specifies the rule.

The evidence packet $\mathcal E_q$ retains attribution and chronological order under a source-token budget $T$:
\begin{equation}
\sum_{e\in\mathcal E_q}|v_e|+\nu_{\rm sep}(\mathcal E_q)\le T.
\label{eq:budget}
\end{equation}
Thus $K$ controls primary seed selection, while $T$ limits delivered source spans and their separators. Packet composition affects evidence use \citep{liu2024lost}; records excluded from this read remain available to later questions.

Transport makes the selected source evidence usable by the current host. The same text can have different token boundaries across models, and their embedding spaces are not shared. MIGRO therefore saves canonical source IDs independently of host input embeddings and constructs receiving-host-compatible inputs at read time. Public embeddings of selected IDs yield weighted lexical hypotheses, which are candidate lexical interpretations of the source pieces. Boundary composition joins subword fragments before target tokenization and embedding lookup \citep{kudo2018sentencepiece}. Appendix~\ref{sec:lexical_transport} gives the equations.

Embedding transport is MIGRO's host input interface; text delivery followed by target tokenization is another possible interface for external memory. Here, continuous embeddings enter through the interface in Section~\ref{sec:portable_objective}. Optional calibration adapts transported evidence or an assistant boundary while leaving the application backbone frozen \citep{li2021prefix,hu2022lora}. Target-context fitting reserves generation space and reduces oversized packets. The retrieval and boundary-composition ablations in Figure~\ref{fig:ablation}a,b test these two stages separately.

\subsection{Reversible Persistence and Model Handoff}
Approximation makes the access structure compact; reversible coding makes the preserved state compact without changing what can be recovered. Address quantization may alter semantic coordinates. Persistence must instead recover the saved coordinates together with the evidence and historical relationships.

A frozen causal predictor estimates token probabilities within bounded pages; it does not rewrite observations. Arithmetic coding turns those probabilities into a reversible representation \citep{witten1987arithmetic,compression}. For predictor-vocabulary token $y_{p,t}$, decoded page-local context $c_{p,t}$, and realized integer-CDF probability $\widetilde p_\phi$, the content length follows
\begin{equation}
\ell_{\rm content}\approx
-\sum_p\sum_{t\in\mathcal T_p}\log_2\widetilde p_\phi(y_{p,t}\mid c_{p,t})
+\ell_{\rm framing}+\ell_{\rm exceptions}.
\label{eq:coding}
\end{equation}
Here $\mathcal T_p$ contains the coded positions in page $p$, and predictive context resets at page boundaries. Surface exceptions retain tokenizer-sensitive content; source-ID exceptions preserve canonical identities when surface reconstruction and retokenization would change them. Addresses, provenance, references, and historical records are archived alongside content.

Every new archive candidate is fully restored and compared with the saved state before acceptance. Selection among exact candidates, including a verified lossless fallback, considers complete encoded bytes and newly required shared dependencies. Appendix~\ref{sec:causal_archive} specifies the coding conditions and selection rule. Figure~\ref{fig:ablation}c evaluates the resulting storage savings.

The restored state remains readable and extendable. A receiving host initializes its selection profile, transport, and optional calibration, then reads inherited records and adds new observations through the same schema. In the running example, both original and revised budgets remain accessible without rebuilding history through the receiver's backbone. Algorithm~\ref{alg:lifecycle} connects recovery to continued writing, whose effect is evaluated in Section~\ref{sec:transfer}.

\subsection{Resource Model and Implementation}
\label{sec:implementation}
A small user state does not by itself determine deployment cost. MIGRO also relies on reusable representation, retrieval, and decoding assets. The relevant questions are how private bytes grow with experience and how fixed resources are shared across independent memories.

Private persistent state contains recoverable evidence, addresses, provenance, references, history, and archival metadata. Shared resources include representation models, tokenizers, retrieval components, host interfaces, and decoding dependencies \citep{agentnative2026}. Temporary workspace holds decoded records, indexes, reconstructed addresses, transport tensors, and inference caches; it is accounted for separately from persistent storage.

Across independent memories $u$, persistent deployment storage is
\begin{equation}
B_{\rm deploy}=B_{\rm shared}+\sum_u B_{{\rm user},u}.
\label{eq:deployment}
\end{equation}
Extending a history adds private bytes; adding independent memories amortizes fixed resources. In the reference configuration, causal prediction uses the common Qwen3-4B checkpoint, which is excluded from the method-specific asset ledger along with the software runtime; any distinct predictor dependency must be charged once. This accounting concerns method resources, rather than a complete machine installation.

Section~\ref{sec:storage} evaluates private-state growth and shared-resource amortization separately. Appendix~\ref{sec:implementation_details} gives configurations and calibration settings, Appendix~\ref{sec:storage_accounting} defines the ledger, and Appendix~\ref{sec:runtime_complexity} describes runtime costs.

\section{Experiments}
\label{sec:experiments}
The experiments ask three questions about the shared experience state: \emph{Can preserved evidence be used effectively?} Quality comparisons and access ablations test whether retained observations become useful answers. \emph{Can experience survive and grow across hosts?} Ordered handoffs and continued writing test inheritance and subsequent accumulation. \emph{Can exact persistence scale economically?} Matched quality--storage measurements, long histories, and shared-resource accounting test the cost of preserving that experience.

Figure~\ref{fig:ablation} connects the three lines through complementary ablations: panels (a,d) examine memory access in Section~\ref{sec:quality}, panels (b,e) examine transfer in Section~\ref{sec:transfer}, and panels (c,f) examine persistence in Section~\ref{sec:storage}.

\subsection{Evaluation Setup}
\label{sec:setup}
We distinguish three execution paths. \emph{Same-host construction and reading} evaluates the complete memory system under one checkpoint. \emph{State Transfer} attaches a receiving host to an accumulated state. \emph{Full-History Rebuilding} reconstructs memory by processing the complete history at the receiver and provides a target-native reference. These paths differ in how history reaches the receiver, even when their final answer scores are similar. Storage comparisons pair quality with complete recoverable states and account separately for shared assets.

LoCoMo, LongMemEval-S, MemOps, MemoryAgentBench, and BEAM-1M cover conversational recall, cross-session reasoning, evolving memory operations, incremental interaction, and long histories \citep{locomo,longmemeval,memops,memoryagentbench,beam}. MemoryAgentBench reports a four-competency macro-average. Baselines include PlugMem, Mem0, TiMem, A-MEM, LSTMem, TransMem, and Doc-to-LoRA \citep{plugmem,mem0,timem,amem,lstmem,transmem,doctolora}. Appendix~\ref{sec:hostgrid} specifies checkpoints; Appendices~\ref{sec:evaluation_details} and~\ref{sec:implementation_details} give scoring, uncertainty estimation, generation budgets, and calibration.

\subsection[Memory Quality and Evidence Organization]{\migroicon\hspace{.38em}Memory Quality and Evidence Organization}
\label{sec:quality}
Can the preserved history support effective answers? MIGRO scores highest on each of the five benchmarks in Table~\ref{tab:mainquality}. Across these tasks, useful evidence may be an individual observation, a revision across sessions, or several records dispersed through a long history. Their common demand is selective access to retained information, beyond keeping that information available.

\begin{table}[!htbp]
\centering
\includegraphics[width=\linewidth]{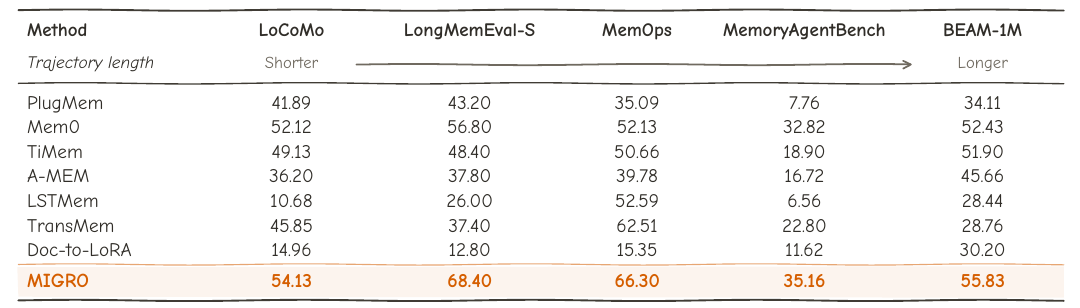}
\caption{Qwen quality under method-native readers and budgets. The arrow spans the trajectory-length progression from LoCoMo through BEAM-1M. LSTMem uses state-only reading on LoCoMo. MemoryAgentBench reports the four-competency macro-average. Metric definitions appear in Table~\ref{tab:benchmarks}.}
\label{tab:mainquality}
\end{table}

\begin{figure}[!htb]
\centering
\includegraphics[width=\linewidth]{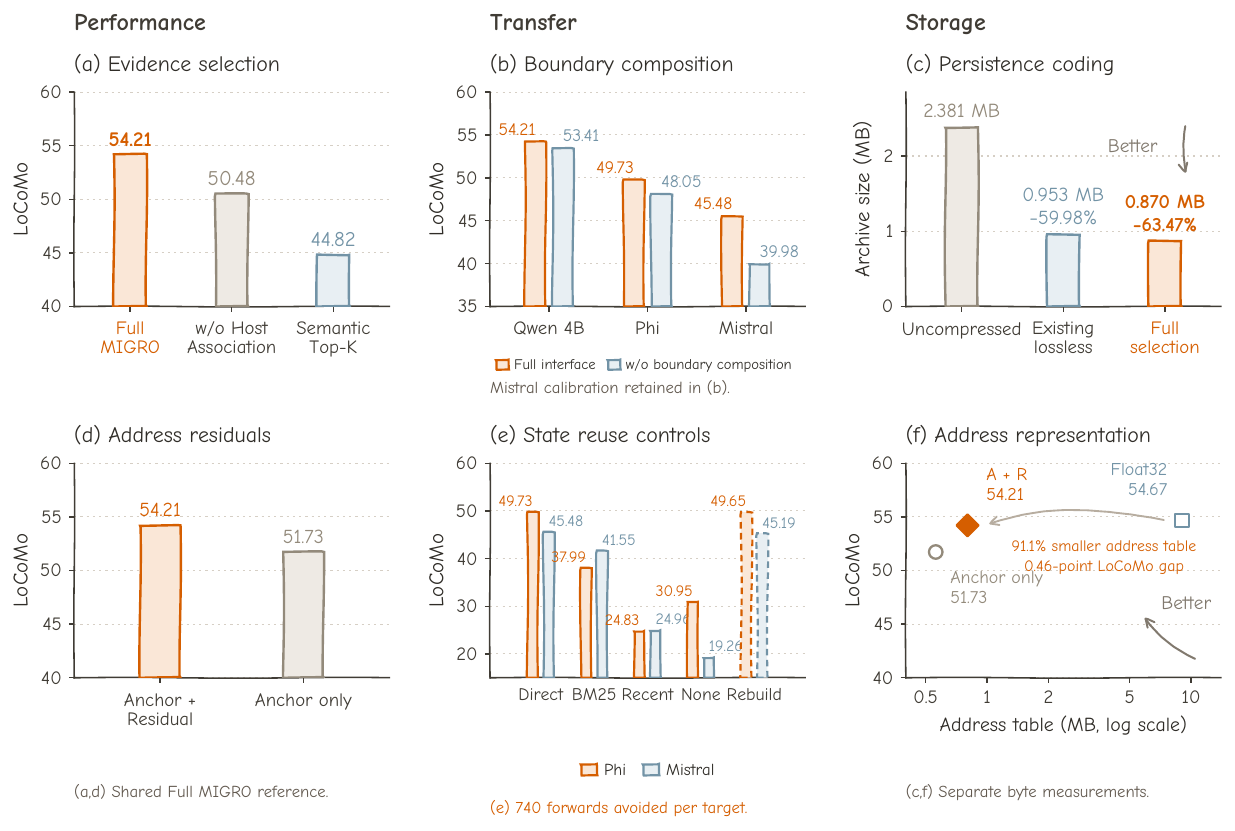}
\caption{Ablations across performance (a,d), transfer (b,e), and storage (c,f). Local ablations use separately evaluated control states with a Qwen reference of 54.21; Table~\ref{tab:mainquality} reports 54.13 for the main-quality evaluation. Higher LoCoMo and lower storage are better; upper left is preferred in (f). Dashed bars in (e) denote target rebuilding; percentages in (c) show savings against the uncompressed archive. A + R denotes Anchor + Residual.}
\label{fig:ablation}
\end{figure}

Relevant evidence is not always the most semantically similar evidence. In Figure~\ref{fig:ablation}a, the full mechanism scores 54.21, removing host-derived contextual associations gives 50.48, and semantic Top-$K$ gives 44.82. Version questions, temporal relations, and reasoning across records can require complementary observations whose value is not captured by similarity alone. The semantic Top-$K$ condition jointly replaces the candidate-search, reranking, and support-selection pipeline, so its gap measures the contribution of this complete evidence-selection path.

Observation-specific placement matters within a shared neighborhood as well. Removing address residuals reduces quality from 54.21 to 51.73 while retaining anchors and source bindings (Figure~\ref{fig:ablation}d). The two ablations identify distinct requirements for memory access: innovations preserve fine-grained semantic placement, while contextual selection determines which observations jointly support a question.

\subsection{Model Capacity and Memory Utilization}
\label{sec:scaling}
\begin{figure}[H]
\centering
\includegraphics[width=.98\linewidth,trim={0bp 11bp 0bp 5bp},clip]{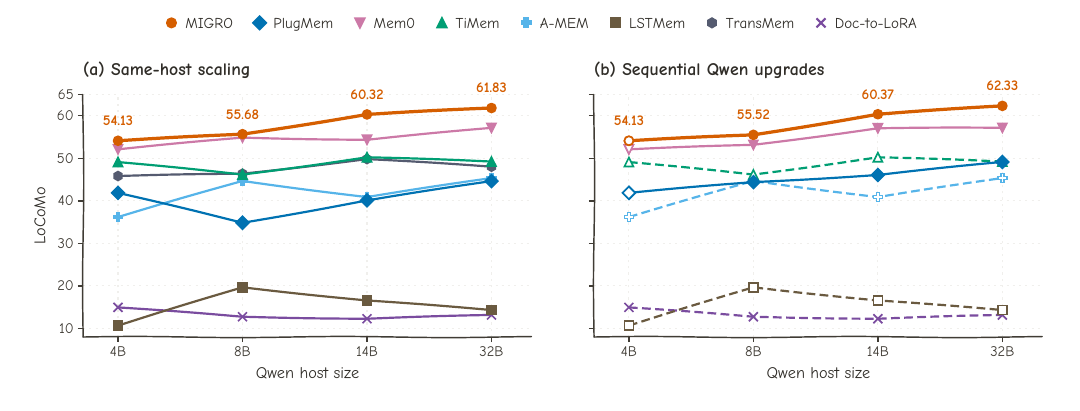}
\caption{LoCoMo quality as Qwen hosts grow. (a) Each host constructs and reads its own memory. (b) Solid curves follow one 4B state through successive upgrades, while dashed curves show full-history rebuilding at each receiver. Both panels use the 4B references from Table~\ref{tab:mainquality}. Tables~\ref{tab:hostscaling} and~\ref{tab:transfer} give the full comparisons.}
\label{fig:scaling}
\end{figure}

When both memory construction and reading move to larger Qwen hosts, MIGRO improves consistently across the evaluated checkpoints, from 54.13 at 4B to 61.83 at 32B (Table~\ref{tab:hostscaling}; Figure~\ref{fig:scaling}). Each checkpoint builds a fresh memory and reads it with the same model. This sweep therefore measures scaling of the complete construction-and-read system.

\begin{table}[!htbp]
\centering\handtablefont
\setlength{\tabcolsep}{3.4pt}
\begin{tabularx}{\linewidth}{@{}Y>{\color{handaccent}}c*{7}{c}@{}}
\toprule
\textbf{Host (build/read)} & \textbf{MIGRO} & \textbf{PlugMem} & \textbf{Mem0} & \textbf{TiMem} & \textbf{A-MEM} & \textbf{LSTMem} & \textbf{TransMem} & \textbf{Doc-to-LoRA}\\
\midrule
Qwen3-4B-Instruct-2507 & \textbf{54.13} & 41.89 & 52.12 & 49.13 & 36.20 & 10.68 & 45.85 & 14.96\\
Qwen3-8B & \textbf{55.68} & 34.86 & 54.88 & 46.19 & 44.70 & 19.70 & 46.39 & 12.77\\
Qwen3-14B & \textbf{60.32} & 40.10 & 54.36 & 50.25 & 40.88 & 16.58 & 49.83 & 12.30\\
Qwen3-32B & \textbf{61.83} & 44.66 & 57.16 & 49.24 & 45.38 & 14.35 & 48.03 & 13.18\\
\bottomrule
\end{tabularx}
\caption{Same-host Qwen LoCoMo quality scaling with the complete baseline set from Table~\ref{tab:mainquality}. Each method constructs a fresh memory for each checkpoint and reads with that same host. The 4B row reuses Table~\ref{tab:mainquality}; target-native rebuilding scores are also used in Table~\ref{tab:transfer} where applicable. LSTMem writes the full history and reads only its persistent state at every size. Bold marks the best available score at each host.}
\label{tab:hostscaling}
\end{table}

Larger hosts do not improve all memory systems proportionally. MIGRO improves at every tier, while LSTMem peaks at 8B under state-only reading. How observations are retained, organized, and presented affects whether greater model capacity translates into better memory answers. These system-level trajectories combine changes in construction and reading; they do not isolate the effect of a stronger reader on a fixed state. Appendix~\ref{sec:implementation_details} records the checkpoint-specific reader and refitting settings.

Improving with a larger host leaves a separate question unresolved: can a stronger or different host inherit the same accumulated state? The next experiment replaces fresh construction with state reuse.

\subsection[Cross-Host Transfer: Experience Beyond Its Original Model]{\migroicon\hspace{.38em}Cross-Host Transfer: Experience Beyond Its Original Model}
\label{sec:transfer}
Portability requires a receiver to use inherited observations and continue accumulating experience. We examine these requirements through ordered Qwen upgrades and two heterogeneous chains, then use matched controls to separate the contributions of memory organization, the receiving interface, and continued writing.

\begin{figure}[!htbp]
\centering
\includegraphics[width=\linewidth]{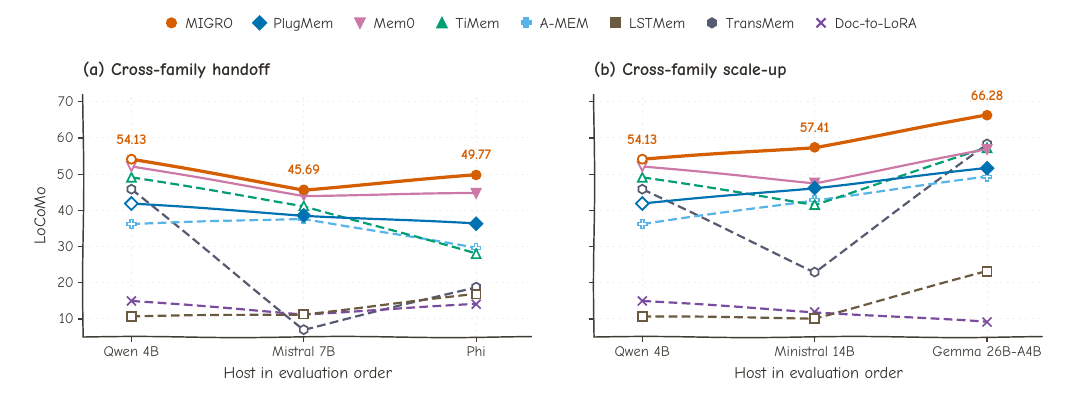}
\caption{LoCoMo quality on two cross-family paths from Qwen 4B. (a) Handoff to Mistral 7B and then Phi. (b) Scale-up to Ministral 14B and then Gemma 26B-A4B. Solid curves show inherited-state transfer and dashed curves show full-history rebuilding. Each method starts at its Table~\ref{tab:mainquality} score.}
\label{fig:transfer}
\end{figure}

\paragraph{Access to inherited experience.}
Table~\ref{tab:transfer} reports receiving-host answer quality across the three chains. Figure~\ref{fig:scaling}(b) tracks Qwen upgrades and Figure~\ref{fig:transfer} shows the two cross-family chains. MIGRO leads the reported method comparisons at each receiving stage. The table's two column groups compare memory lifecycles as well as answer quality: MIGRO, PlugMem, and Mem0 transfer existing states; TiMem, A-MEM, LSTMem, TransMem, and Doc-to-LoRA rebuild from the full history at each receiver. A useful transferred state provides evidence without repeating that historical construction through the new backbone.

\begin{table}[!htbp]
\centering\handtablefont
\setlength{\tabcolsep}{2.8pt}
\begin{tabularx}{\linewidth}{@{}Y>{\color{handaccent}}c*{7}{c}@{}}
\toprule
& \multicolumn{3}{c}{State Transfer} & \multicolumn{5}{c}{Full-History Rebuilding}\\
\cmidrule(lr){2-4}\cmidrule(l){5-9}
\textbf{Stage / path} & \textbf{MIGRO} & \textbf{PlugMem} & \textbf{Mem0} & \textbf{TiMem} & \textbf{A-MEM} & \textbf{LSTMem} & \textbf{TransMem} & \textbf{Doc-to-LoRA}\\
\midrule
\multicolumn{9}{@{}l}{\textit{LoCoMo Qwen chain: 4B training/source stage, then 8B, 14B, and 32B}}\\[2pt]
4B$\to$8B & \textbf{55.52} & 44.41 & 53.18 & 46.19 & 44.70 & 19.70 & 46.39 & 12.77\\
4B$\to$8B$\to$14B & \textbf{60.37} & 46.10 & 57.07 & 50.25 & 40.88 & 16.58 & 49.83 & 12.30\\
4B$\to$8B$\to$14B$\to$32B & \textbf{62.33} & 49.15 & 57.17 & 49.24 & 45.38 & 14.35 & 48.03 & 13.18\\[3pt]
\midrule
\multicolumn{9}{@{}l}{\textit{Heterogeneous chain: Qwen 4B source, then Mistral 7B and Phi}}\\[2pt]
Qwen 4B$\to$Mistral 7B & \textbf{45.69} & 38.60 & 44.06 & 41.22 & 37.73 & 11.31 & 7.04 & 11.32\\
Qwen 4B$\to$Mistral 7B$\to$Phi & \textbf{49.77} & 36.30 & 44.72 & 28.03 & 29.57 & 16.84 & 18.79 & 14.07\\[3pt]
\midrule
\multicolumn{9}{@{}l}{\textit{Heterogeneous scale-up: Qwen 4B $\to$ Ministral 14B $\to$ Gemma 26B-A4B}}\\[2pt]
Qwen 4B$\to$Ministral 14B & \textbf{57.41} & 46.19 & 47.55 & 41.60 & 42.84 & 10.18 & 22.94 & 11.92\\
4B$\to$Ministral 14B$\to$Gemma 26B-A4B & \textbf{66.28} & 51.63 & 56.83 & 57.29 & 49.26 & 23.16 & 58.36 & 9.15\\
\bottomrule
\end{tabularx}
\caption{LoCoMo migration QA by execution mode. MIGRO, PlugMem and Mem0 use State Transfer; TiMem, A-MEM, LSTMem, TransMem and Doc-to-LoRA use Full-History Rebuilding at each receiver. TransMem's Qwen references come from Table~\ref{tab:hostscaling}. Doc-to-LoRA's Ministral and Gemma results each cover all 1,986 questions. Appendix~\ref{sec:hostgrid} gives exact checkpoints. Bold marks the highest score in rows with method comparisons.}
\label{tab:transfer}
\end{table}

\paragraph{Heterogeneous scale-up.}
MIGRO carries the Qwen 4B memory state first to Ministral 14B and then to Gemma 26B-A4B. LoCoMo reaches 57.41 at the first receiving stage and 66.28 at the second, an increase of 8.87 points along the inherited-state path. The closest measured alternatives are transferred Mem0 at Ministral (47.55) and TransMem rebuilt from full history at Gemma (58.36). MIGRO leads these references by 9.86 and 7.92 points, respectively.

\paragraph{Value beyond carrying a history.}
The core transfer controls ask whether inherited organization contributes beyond having historical text available (Figure~\ref{fig:ablation}e; Appendix~\ref{sec:evaluation_details}, Figure~\ref{fig:coretransfer}). The full mechanism outperforms BM25, recent-history windows, and no-memory access under every evaluated host. Against BM25, the reported paired gains are 11.74 points on Phi and 3.93 on Mistral. Retention supplies the observations, but their organization and access determine how much useful experience reaches the receiver.

Direct reuse also approaches target-side rebuilding quality: 49.73 versus 49.65 on Phi, and 45.48 versus 45.19 on Mistral. Each receiver avoids 740 memory-building model forwards while still performing normal answering. This comparison identifies the operational value of the shared state: the receiver can use accumulated experience at near-rebuilding quality without reconstructing its history. Paired intervals and the complete controls appear in Appendix~\ref{sec:evaluation_details}.

\paragraph{Compatibility across token boundaries.}
The interface ablation tests a more local obstacle to inheritance (Figure~\ref{fig:ablation}b). Removing boundary composition lowers quality under all three hosts. On Mistral, with calibration retained, quality falls from 45.48 to 39.98. Cross-model compatibility therefore involves more than matching embedding widths: source subword fragments must be composed into lexical representations that the receiver can interpret effectively. The matched controls connect this component of the transport interface to answer quality without changing the persistent evidence.

\begin{figure}[!htbp]
\centering
\includegraphics[width=\linewidth]{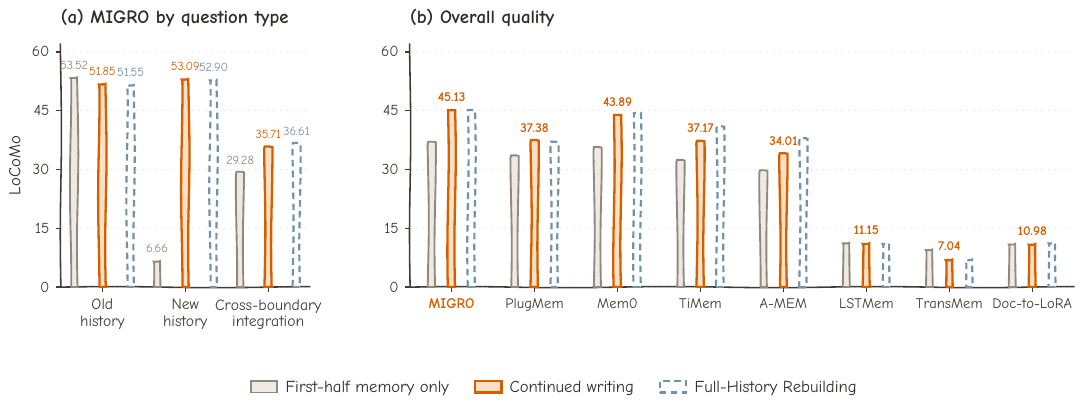}
\caption{Continuity of memory evolution after handoff. (a) MIGRO's quality on inherited, new-history, and cross-boundary questions. (b) Overall quality under three memory conditions; numbers label continued-writing scores. The continuation operations and source-state pairing are detailed in Appendix~\ref{sec:joint_details}.}
\label{fig:continuedwrite}
\end{figure}

\paragraph{Continued growth after handoff.}
A portable memory should continue evolving under its receiving host. Figure~\ref{fig:continuedwrite} shows that extending the inherited state improves access to post-handoff observations and cross-boundary evidence while retaining effective access to earlier history. Overall quality reaches 45.13, close to full-history rebuilding at 45.19, without reconstructing the pre-handoff history through the receiver. Mem0, PlugMem, A-MEM, and TiMem also benefit from continuation, while MIGRO attains the highest continued-write score in Figure~\ref{fig:continuedwrite}b. LSTMem, Doc-to-LoRA, and TransMem show no gain in their overall scores when extending their first-half memories. Portability thus includes continuity of experience accumulation: model replacement changes how history is interpreted and extended while the inherited state remains the basis for future reasoning.

\subsection[Storage Efficiency: Fidelity Without Redundant Persistence]{\migroicon\hspace{.38em}Storage Efficiency: Fidelity Without Redundant Persistence}
\label{sec:storage}
Storage efficiency must explain both why a state is compact and whether it can remain economical as experience grows. We connect representation and coding ablations to matched-state quality, marginal history growth, and deployment-scale sharing.

\paragraph{Where the byte savings come from.}
Figure~\ref{fig:ablation}c separates the basic gain from lossless coding and the additional gain from archive selection. The uncompressed archive occupies 2.381 MB; existing lossless coding reduces it to 0.953 MB; full selection reduces it to 0.870 MB. Full selection saves about 63.5\% against the uncompressed archive and a further 8.7\% against existing lossless coding. Exact recovery is retained in every accepted encoded candidate.

Address organization provides a separate saving (Figure~\ref{fig:ablation}f). Anchor + Residual reduces address-table bytes by 91.1\% relative to float32, with a 0.46-point LoCoMo gap. The representation used to locate evidence can therefore be substantially compressed without imposing the same loss on the evidence itself. These address conditions also supply the quality ablation in panel (d); the address comparison counts semantic addresses, while the coding comparison counts the complete archive. Appendix~\ref{sec:storage_accounting} gives their exact sizes.

\begin{figure}[!htbp]
\centering
\includegraphics[width=\linewidth]{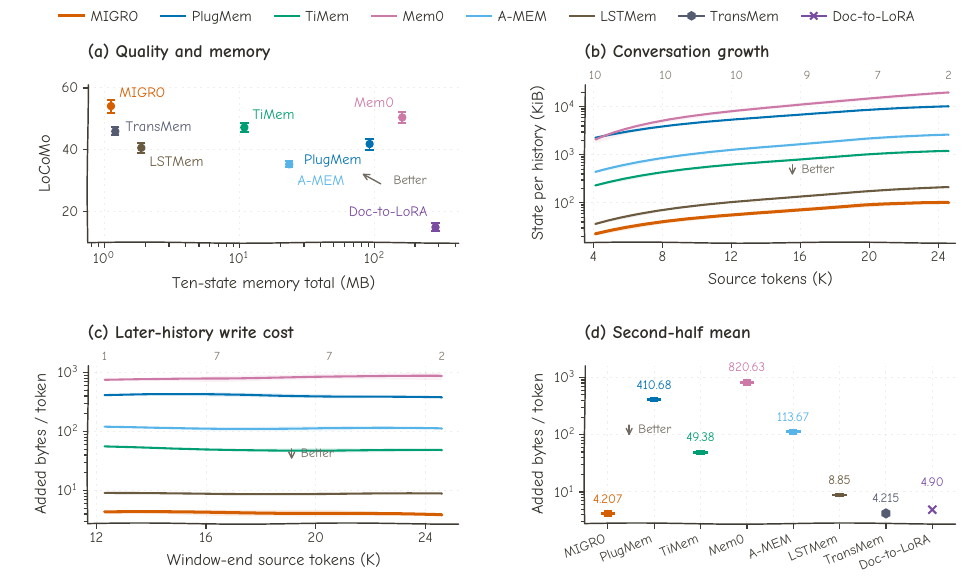}
\caption{LoCoMo quality and storage: (a) quality versus matched-state bytes; (b) state growth; (c) later-history marginal cost; (d) second-half mean. TransMem and Doc-to-LoRA summary measurements enter (a,d). Counts above (b,c) identify eligible conversations; available bands and bars are 95\% bootstrap intervals. Fixed assets are accounted for separately.}
\label{fig:storagequality}
\end{figure}

\paragraph{Useful answers from compact states.}
The matched-state comparison asks whether compactness coexists with answer quality in the same memory (Table~\ref{tab:storage}; Figure~\ref{fig:storagequality}). MIGRO scores 54.13 with 1.115 MB across ten conversation states. TransMem uses 1.200 MB and scores 45.96: MIGRO gains 8.17 points while using 7.1\% less storage. Mem0 scores 50.45 at 160.671 MB. Every point pairs the quality and bytes of the actual evaluated state, exposing the trade-off between short-history compactness and effective access to retained evidence. Doc-to-LoRA further separates small write increments from small states: its 4.9 bytes/token LoCoMo late-half rate coexists with 283.14 MB of accumulated private state.

\begin{table}[!htbp]
\centering\handtablefont
\begin{tabularx}{\linewidth}{@{}L{.15\linewidth}*{3}{>{\centering\arraybackslash}X}@{}}
\toprule
\textbf{Method} & \textbf{\shortstack{LoCoMo $\uparrow$\\(0--100)}} & \textbf{95\% CI} & \textbf{\shortstack{Native memory $\downarrow$\\(MB, ten states)}}\\
\midrule
PlugMem & 41.89 & [40.01, 43.39] & 92.078\\
Mem0 & 50.45 & [48.81, 52.18] & 160.671\\
TiMem & 47.22 & [45.70, 48.56] & 10.851\\
A-MEM & 35.29 & [34.46, 36.33] & 23.340\\
LSTMem & 40.64 & [39.11, 42.10] & 1.871\\
TransMem & 45.96 & [44.80, 47.26] & 1.200\\
Doc-to-LoRA & 14.96 & [13.83, 16.24] & 283.140\\
\rowcolor{handwash} \textcolor{handaccent}{\textbf{MIGRO}} & \textcolor{handaccent}{\textbf{54.13}} & \textcolor{handaccent}{[51.93, 56.22]} & \textcolor{handaccent}{\textbf{1.115}}\\
\bottomrule
\end{tabularx}
\caption{LoCoMo quality and native storage from the same evaluated states, under the Qwen reference host and each method's native reader. These storage replicas are evaluated separately from Table~\ref{tab:mainquality}. LSTMem uses the retained-history reader for this paired quality--storage measurement. Storage is summed across ten conversations; intervals cluster by conversation.}
\label{tab:storage}
\end{table}

\begin{figure}[!htbp]
\centering
\includegraphics[width=\linewidth]{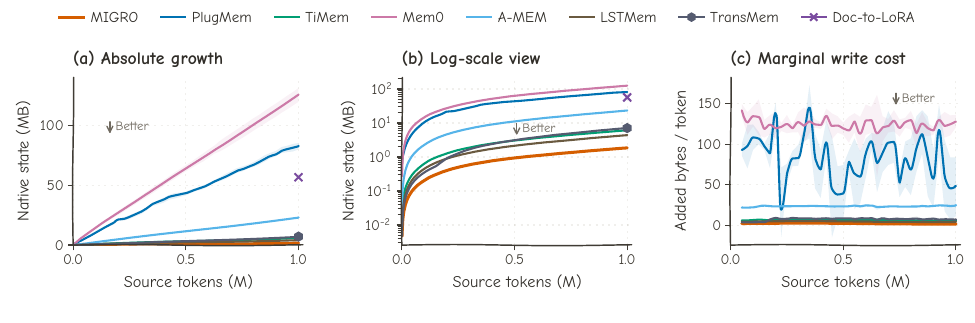}
\caption{Million-token BEAM storage: (a) absolute state growth; (b) the same measurements on a logarithmic axis; (c) added bytes per token over successive windows. Curves show complete prefix series; the Doc-to-LoRA marker shows its measured million-token endpoint; late-half averages appear in Table~\ref{tab:growth}. Available bands are 95\% history-bootstrap intervals.}
\label{fig:storagegrowth}
\end{figure}

\paragraph{Economical growth within a history.}
A small snapshot matters only if the memory can continue growing economically. The LoCoMo prefixes in Figure~\ref{fig:storagequality} and million-token BEAM histories in Figure~\ref{fig:storagegrowth} examine this growth directly. Across the five BEAM histories, MIGRO averages 1.857 MB per million-token history and adds 1.809 bytes per token over the second half (Table~\ref{tab:growth}). In the million-token comparison, TransMem's BEAM state occupies 7.092 MB, with a late-half rate of 7.778 bytes/token. MIGRO's lower marginal cost keeps its long-history state compact, widening the absolute storage advantage as experience accumulates.

\begin{table}[!htbp]
\centering\handtablefont
\begin{tabularx}{\linewidth}{@{}L{.14\linewidth}*{4}{>{\centering\arraybackslash}X}@{}}
\toprule
\textbf{Method} & \textbf{\shortstack{State at 1M\\MB/history $\downarrow$}} & \textbf{95\% CI (MB)} & \textbf{\shortstack{500K--1M\\bytes/token $\downarrow$}} & \textbf{95\% CI}\\
\midrule
PlugMem & 82.871 & [79.781, 85.277] & 79.608 & [76.162, 84.012]\\
Mem0 & 125.499 & [119.613, 131.087] & 123.093 & [115.187, 130.500]\\
TiMem & 6.166 & [6.119, 6.198] & 6.178 & [6.114, 6.225]\\
A-MEM & 23.191 & [22.914, 23.496] & 23.639 & [23.188, 24.073]\\
LSTMem & 4.371 & [4.276, 4.437] & 4.370 & [4.233, 4.453]\\
TransMem & 7.092 & [7.018, 7.154] & 7.778 & [7.709, 7.854]\\
Doc-to-LoRA & 56.625 & [56.625, 56.625] & 28.313 & [28.308, 28.319]\\
\rowcolor{handwash} \textcolor{handaccent}{\textbf{MIGRO}} & \textcolor{handaccent}{\textbf{1.857}} & \textcolor{handaccent}{[1.818, 1.886]} & \textcolor{handaccent}{\textbf{1.809}} & \textcolor{handaccent}{[1.732, 1.865]}\\
\bottomrule
\end{tabularx}
\caption{Million-token state sizes and 500K--1M marginal cost. Means and 95\% history-bootstrap intervals use the same five paired histories for every method; rates use actual source-token boundaries.}
\label{tab:growth}
\end{table}

\begin{figure}[!htbp]
\centering
\includegraphics[width=\linewidth]{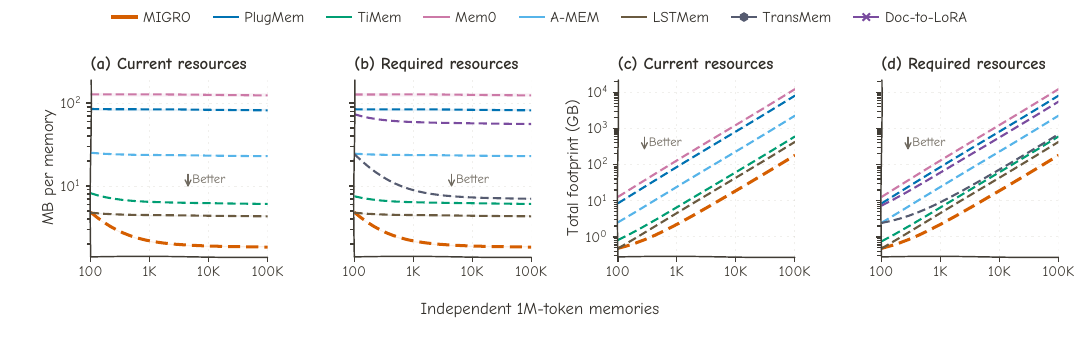}
\caption{Shared-resource estimates for independent million-token memories. Panels (a,b) show average MB per memory, and (c,d) show total method-system footprint. Panels (a,c) count current deployed resources, and (b,d) count required resources, including estimates for TransMem and Doc-to-LoRA. Each method shares one fixed asset bundle. Available bands propagate state-mean intervals.}
\label{fig:sharedstorage}
\end{figure}

\Needspace{7\baselineskip}
\paragraph{Sharing fixed resources across histories.}
The deployment estimates expose a second scale of economy (Table~\ref{tab:sharedstorage}; Figure~\ref{fig:sharedstorage}). With 100 independent million-token memories, LSTMem has a slightly lower estimated average cost: 4.775 MB versus MIGRO's 4.829 MB. With 1,000 memories, MIGRO falls to 2.154 MB per memory, compared with 4.411 MB for LSTMem and 8.812 MB for TransMem. TransMem's required shared bundle is 1.720 GB. Wider sharing amortizes MIGRO's fixed resources, making its smaller private state increasingly important.

\begin{table}[!htbp]
\centering\handtablefont
\begin{tabularx}{\linewidth}{@{}L{.16\linewidth}*{4}{>{\centering\arraybackslash}X}@{}}
\toprule
& \multicolumn{4}{c}{Independent million-token memories}\\
\cmidrule(l){2-5}
\textbf{Method} & \textbf{100} & \textbf{1,000} & \textbf{10,000} & \textbf{100,000}\\
\midrule
PlugMem & 84.237 & 83.008 & 82.885 & 82.873\\
Mem0 & 127.018 & 125.651 & 125.514 & 125.500\\
TiMem & 7.532 & 6.303 & 6.180 & 6.167\\
A-MEM & 24.557 & 23.328 & 23.205 & 23.193\\
LSTMem & \textbf{4.775} & 4.411 & 4.375 & 4.371\\
TransMem & 24.292 & 8.812 & 7.264 & 7.109\\
Doc-to-LoRA & 73.465 & 58.309 & 56.794 & 56.642\\
\rowcolor{handwash} \textcolor{handaccent}{\textbf{MIGRO}} & \textcolor{handaccent}{4.829} & \textcolor{handaccent}{\textbf{2.154}} & \textcolor{handaccent}{\textbf{1.887}} & \textcolor{handaccent}{\textbf{1.860}}\\
\bottomrule
\end{tabularx}
\caption{Estimated average storage in MB per independent million-token memory, using required method resources shared once. The private-state term uses the reported million-token endpoints.}
\label{tab:sharedstorage}
\end{table}

Within a history, economy depends on marginal bytes per observation. Across histories, it also depends on how widely representation, retrieval, and decoding assets are shared. The estimates combine measured state sizes with required shared resources under the accounting model in Section~\ref{sec:implementation}; they describe method-system storage. Appendix~\ref{sec:storage_accounting} gives the complete definitions.

\FloatBarrier
\subsection{Joint Design Implications}
\label{sec:joint}
Evaluating long-term memory requires following experience across reads, writes, and host changes. Memory quality depends on whether an access interface exposes useful relationships among retained observations. Portability includes the ability to extend inherited experience under a new host. Persistent cost has two scales: marginal growth within one history and infrastructure shared across many histories. Evaluating one snapshot or one receiving-host score misses these lifecycle properties.

Figure~\ref{fig:radar}a highlights six selected performance strengths. The largest gains are on LongMemEval-S multi-session QA and temporal reasoning, where MIGRO leads the strongest displayed baselines by 15.79 and 18.80 points, respectively. The transfer panel shows quality across receiving hosts, memory continuity and growth, and reuse without historical replay. Its portable-efficiency axis pairs continued-writing quality with the receiver's private state size. The storage panel connects compact private states with economical growth: MIGRO has the lowest displayed private footprints, marginal costs over the matched half-history windows, and BEAM deployment cost, together with the highest LoCoMo score per private MB. Appendix~\ref{sec:joint_details} defines the profile axes and normalization. The joint frontier in Figure~\ref{fig:frontier} places continued-writing quality, rebuilding-relative quality and paired private cost in one Mistral setting. MIGRO combines the highest inherited-state answer quality with the smallest inherited-state footprint, yielding 2.60 times the best baseline's answer quality per private MB across all eight methods.

\subsection{What Remains Difficult in Long-Term Memory}
\label{sec:failures}
On BEAM-1M, information extraction scores 66.92 and preference following 79.64, compared with 39.64 for contradiction resolution and 37.44 for event ordering (Appendix~\ref{sec:evaluation_details}, Table~\ref{tab:failures}). Preserving facts is different from reconstructing their relationships. The harder questions require the reader to connect versions, recover temporal order, and combine evidence across records. Their scores do not identify one failing stage, but they expose the distance between retained observations and an accessible historical account.

Exact preservation keeps historical distinctions available. The remaining challenge is to make the relationships among those distinctions accessible when a question demands them. By keeping experience independent of its current reader, MIGRO provides a persistent foundation on which future memory access and reasoning can improve.

\FloatBarrier
\Needspace{8\baselineskip}
\section{Conclusion}
\label{sec:conclusion}
MIGRO organizes long-term memory around the continuity of experience across model changes. By separating preservation from access, it keeps historical evidence exact while allowing semantic organization and query-specific inputs to adapt. The completed evaluations connect this design to competitive question answering, effective state reuse across heterogeneous hosts, and continued writing close to full-history rebuilding, alongside compact persistence and low marginal storage growth. Quality, portability, and storage jointly determine whether accumulated experience remains useful as both the history and its consuming model change.

The broader implication is a different relationship between model progress and memory growth. A model upgrade can change how earlier observations are interpreted without erasing the evidence available to future questions. Continued writing extends that history under the receiving host, making portability a continuity of accumulation rather than a one-time transfer. Preserving experience independently of its current reader leaves room for advances in retrieval and historical reasoning to improve how that experience is used. The model can change; the history that supports future reasoning can continue to grow.

\Needspace{8\baselineskip}
\begingroup
\fontsize{9}{11}\selectfont
\setlength{\bibsep}{1.5pt plus.5pt}
\bibliographystyle{unsrtnat}
\bibliography{references,references_foundations,references_recent}
\endgroup
\Needspace{12\baselineskip}
\appendix
\section{Relationship to Prior Work}
\label{sec:prior}
MIGRO connects four research threads: learned access to retained representations, external evidence separated from active context, hybrid representations with explicit preservation criteria, and interfaces that reuse memory across consumers. The distinctions concern what is stored, how it is accessed and which resources remain tied to a model.

\subsection{Neural and latent memory: extending computational state}
Memory Networks make addressable memory participate in inference; the end-to-end formulation learns repeated attention over supporting memories \citep{weston2015memory,sukhbaatar2015memory}. Compressive Transformers extend the past through compressed activations, while Memorizing Transformers retrieve internal key--value pairs through approximate nearest-neighbor lookup \citep{rae2020compressive,wu2022memorizing}. In both cases, the retained representation extends what the model can use beyond its ordinary active window, under conventions designed with the consuming architecture.

Factual model editing changes selected weights to alter stored associations \citep{meng2022rome}. MemoryLLM, LSTMem and TransMem explore latent pools, online state accumulation or context-derived interventions as further forms of learned memory \citep{memoryllm,lstmem,transmem}. Their storage and update rules give retained experience different computational roles. MIGRO uses learned representations for semantic organization and interpretation, while independent source bindings define historical identity.

Metis learns native memory procedures through memory-specific mid-training. Its persistent dynamic state is updated by forward computation and accessed through memory attention \citep{metis2026}. MemT-1.0 extends this approach with dynamic local memory, trainable hyper memory, stable token-level Gated DeltaNet-2 updates and Mixture-of-Storage capacity scaling \citep{memt2026}. $\delta$-mem instead adds a compact associative matrix alongside a frozen full-attention backbone, writes with a gated delta rule, and uses the readout for low-rank attention corrections \citep{deltamem2026}. These designs make compact retained state part of active computation. MIGRO assigns a separate role to its growing canonical history: preserving exact observations that later computations can revisit.

MemOS broadens the managed objects to plaintext, activation and parameter memory. Its MemCubes include provenance and versions and support composition, migration and lifecycle management \citep{memos2025}. This system-level view establishes that portable memory can encompass several representation types. MIGRO specifies which observation fields persist, how each host reads and extends them, and how an archive recovers the saved state.

\subsection{Explicit and retrieval-augmented memory: separating records from active context}
REALM and RAG couple neural models to externally retrieved documents, and dense passage retrieval provides learned semantic access to that corpus \citep{guu2020realm,lewis2020rag,karpukhin2020dpr}. The corpus can persist independently of the model's parameters while retrieval constructs a query-specific input. This is an established foundation for memory surviving a model instance.

Agent memory extends this separation to evolving interactions. Managed recall, reflections, linked descriptors, extracted knowledge and temporal hierarchies organize an agent's experience \citep{memgpt,agents,amem,mem0,timem,plugmem}. Several such systems already retain original content or pursue model independence. The relevant comparison is therefore how retention, bounded access, continued growth and host replacement are connected, rather than whether a system uses external storage.

Recent work exposes the lifecycle structure more directly. Hindsight separates world facts, experiences, entity summaries and evolving beliefs, while EverMemOS connects episodic MemCells, consolidated MemScenes and reconstructive recall \citep{hindsight2025,evermemos2026}. Zep uses time-aware facts, and MemoryLACE retains provenance-bearing textual units linked by merge, supersession and contradiction relations \citep{zep2025,memorylace2026}. These structures make historical relationships part of retrieval rather than leaving them implicit in an undifferentiated transcript.

Other designs change the granularity or timing of access. H-Mem combines temporal and semantic trees, and HiGMem uses event-summary anchors to reach individual turns \citep{hmem2026,higmem2026}. MRAgent explores a Cue--Tag--Content graph during reasoning, while RecMem delays consolidation until recurrent evidence supports it and refines access to finer facts \citep{mragent2026,recmem2026}. Such mechanisms determine which retained evidence can be recovered economically. Long-context position effects further distinguish available records from evidence a model can use effectively \citep{liu2024lost}. MIGRO makes the delivered packet and its source-token cost explicit at this boundary.

\subsection{Hybrid representation and evidence preservation: assigning fidelity}
Product and additive quantization compress vector coordinates for approximate search \citep{jegou2011pq,babenko2014aq}. An approximate address can locate an exact object, which motivates assigning separate fidelity criteria to search geometry and source evidence. Database provenance and temporal databases provide the corresponding distinction between a current value, its origin and its historical states \citep{cheney2009provenance,snodgrass1999temporal}. MIGRO combines quantized address organization with explicit source attribution, record order and retained content.

Mem++ is a close preservation strategy: it keeps complete dated, attributed documents and selects relevant versions at read time \citep{memplus2026}. SimpleMem instead synthesizes compact semantic units and plans retrieval by intent; VikingMem evolves an event/entity base with topic-wise temporal compression \citep{simplemem2026,vikingmem2026}. The preservation criterion matters to this comparison. MIGRO's archive recovers the saved source IDs, coordinates and bookkeeping, while its active packet is selective.

DeMem, \textit{Remember the Decision, Not the Description}, measures compression through achievable decision quality under a fixed number of runtime states. Certified conflicts trigger partition refinement \citep{demem2026}. This is complementary to retaining canonical experience for an expanding family of future questions: runtime abstraction limits what one decision process carries, whereas durable preservation determines what future processes can revisit.

Information theory distinguishes relevance-preserving abstraction from redundancy removed by a reversible code \citep{tishby1999bottleneck,cover2006information,witten1987arithmetic,compression}. Section~\ref{sec:theory} expresses preservation through query-distinguishable histories. The causal archive then exploits statistical regularity without changing that saved history. Semantic quantization changes address coordinates, selection limits the evidence packet, and lossless coding reduces archive bytes while preserving the saved state.

\subsection{Portable interfaces: changing consumers around retained experience}
A persistent record and a model input have different lexical conventions. Token identities, segmentation and normalization depend on the tokenizer \citep{kudo2018sentencepiece}. MIGRO separates a canonical source vocabulary from the receiving vocabulary and constructs target embeddings through lexical/subword hypotheses. Exact source recovery establishes what survived. Successful host interpretation establishes whether the receiver can use it.

Frozen-memory extraction studies this receiving boundary by attaching an external learned table to another backbone through compatible or adapted readers \citep{frozenmemory2026}. Model-upgrade experiments distinguish construction losses from retrieval and interpretation failures, including repair enabled by original histories \citep{modelupgrade2026}. Metis also applies its architecture and mid-training recipe to compatible Llama and Gemma backbones \citep{metis2026}. That comparison tests architectural and training transfer. MIGRO tests a complementary object: the same accumulated user state handed between receivers, with continued writes and exact saved-state restoration.

Continuous prefixes and LoRA localize adaptation while keeping pretrained backbone weights frozen \citep{li2021prefix,hu2022lora}. LongMem decouples a frozen backbone from a learned SideNet reader, while TransMem and Doc-to-LoRA derive reusable latent interventions or low-rank adapters from context \citep{longmem,transmem,doctolora}. These designs show several ways to place a reusable memory or conditioning interface around a host. In MIGRO, optional calibration is a shared access asset rather than per-user experience. \label{sec:doctolora_setting}The evaluated Doc-to-LoRA baseline uses its stated stabilized native-window $1/N$ configuration.

MemCollab addresses another form of transfer: contrasting model-generated reasoning trajectories to derive common task-level constraints \citep{memcollab2026}. Its reasoning abstractions concern a different retained object from MIGRO's observations and source identities. Agent-native evaluations and phase-aware workload studies connect these representation choices to maintenance, construction, retrieval and generation cost \citep{agentnative2026,memoryworkloads2026}. MIGRO's accounting isolates complete private bytes, marginal history growth and shared persistent resources within that broader cost picture.

Table~\ref{tab:prior} locates selected recent mechanisms by their retained objects, consuming interfaces, growth operations and preservation criteria.

\begin{table}[!htbp]
\centering\handtablefont\footnotesize
\setlength{\tabcolsep}{4pt}
\begin{tabularx}{\linewidth}{@{}L{.16\linewidth}YYY@{}}
\toprule
\textbf{Work} & \textbf{Evidence / organization} & \textbf{Host boundary / growth} & \textbf{Preservation / cost object}\\
\midrule
Mem++ \citep{memplus2026} & Whole dated documents; lexical--semantic retrieval. & Document consumption; append later versions. & Original documents; generative write calls.\\[3pt]
MemOS \citep{memos2025} & Typed MemCubes with provenance and versions. & MemCube migration, composition and lifecycle evolution. & Content/metadata management; heterogeneous memory resources.\\[3pt]
Frozen-memory extraction \citep{frozenmemory2026} & External learned table and common addressing. & Frozen-source reuse with compatible or adapted target readers. & Reused learned artifact; reader adaptation and QA.\\[3pt]
Model-upgrade study \citep{modelupgrade2026} & Raw histories, retrieval chunks, notes and fixed-schema graphs. & Direction-specific writer/reader swaps and source-assisted repair. & Construction/access diagnostics; QA continuity and repair.\\[3pt]
MemoryLACE \citep{memorylace2026} & Atomic text and provenance; sparse lifecycle relations. & Evidence access across QA backbones; merge, supersession and contradiction links. & Relation-aware evidence units; end-to-end runtime.\\[3pt]
VikingMem \citep{vikingmem2026} & Event/entity base with topic-wise timelines. & Application-level memory base; event-driven state evolution. & Temporal abstraction and compression; recall and latency.\\[3pt]
DeMem \citep{demem2026} & Decision-oriented partition of history--query contexts. & Decision readout; certified partition refinement. & Decision distortion; $K$ runtime memory states.\\[3pt]
\rowcolor{handwash} \textcolor{handaccent}{\textbf{MIGRO}} & Exact source bindings; quantized anchors and innovations. & Receiving-host input interface; append after handoff. & Exact saved-state recovery; private and shared persistent bytes.\\
\bottomrule
\end{tabularx}
\caption{Selected recent mechanisms across MIGRO's lifecycle boundaries. Paired columns identify the retained object and access structure, the consuming interface and growth operation, and the preservation criterion and resource object.}
\label{tab:prior}
\end{table}

\FloatBarrier

\FloatBarrier
\Needspace{8\baselineskip}
\section{Unified Notation}
Source, coding and target tokenizers define different sequence lengths. Table~\ref{tab:notation} distinguishes their symbols alongside the state, access and storage quantities. The host profile specifies calibration and task budgets; the private state retains experience.
\begingroup
\handtablefont
\fontsize{9}{10.5}\selectfont
\setlength{\LTpre}{6pt}
\setlength{\LTpost}{8pt}
\begin{longtable}{@{}L{.21\linewidth}L{\dimexpr.79\linewidth-2\tabcolsep\relax}@{}}
\caption{Unified symbols. Address approximation occurs at writing; exact archival recovery is defined over the saved representation.}
\label{tab:notation}\\
\toprule
\textbf{Symbol} & \textbf{Meaning}\\
\midrule
\endfirsthead
\toprule
\textbf{Symbol} & \textbf{Meaning}\\
\midrule
\endhead
\bottomrule
\endfoot
\endlastfoot
$\mathcal M_n$ & Logical persistent state after $n$ accepted observations.\\
$o_i=(x_i,m_i)$ & Observation content and its available provenance.\\
$h_i,\ h,\ \theta_h$ & Writing host, reading host, and frozen host parameters.\\
$\mathcal A,\mathcal D,\mathcal X,\mathcal P,\mathcal G,\mathcal V$ & Anchors, address innovations, exact content, provenance, reference links, and retained history.\\
$\sigma,\tau_s,v_i,E_s$ & Canonicalization, shared source tokenizer, exact ordered source IDs, and public source embedding table.\\
$k_i,u_i,\mu,U$ & Public semantic address, shared coordinates, center, and basis.\\
$A_j,a_i,W_\delta,\widehat\delta_i$ & Shared anchor, assigned anchor index, innovation subspace, and saved innovation.\\
$\mathcal W_i,\rho_i$ & Bounded same-session past window and chosen predecessor relation.\\
$\mathcal E_q,K,T$ & Delivered evidence, relevance-seed count, and source-token budget.\\
$T_h,\mathcal I_h,\eta_h$ & Lexical/subword transport, host assembly operator, and optional calibrated residual parameters.\\
$\Omega_g,\tau_h,E_h$ & Composed lexical hypotheses, target tokenizer, and target embedding table.\\
$y_{p,t},c_{p,t},\widetilde p_\phi$ & Coding-vocabulary token, page-local causal context, and realized integer-CDF probability.\\
$f_p,\mathcal T_p,b_p$ & Page framing, actually coded positions, and arithmetic bitstream.\\
$\mathcal B,\Pi_h$ & Shared dependency bundle and declared host access profile.\\ 
$S_n$ & Snapshot whose exact decoding restores the saved $\mathcal M_n$.\\
$L,B_{\rm user},\beta$ & Original source-body token count, complete private persistent bytes, and incremental bytes/token.\\
$N,B_{\rm shared},C_{\rm memory}(N)$ & Number of independent memories, fixed method assets, and mean storage per memory after sharing.\\
$R_n,\bar m_i,\mathcal Q,\sim_{\mathcal Q}$ & Canonical evidence history, saved contextual fields, future question family, and equivalence under its evidence-determined answers.\\
$F,\operatorname{Ans}(q,R)$ & Query-independent abstract encoding and answer determined by history $R$.\\
$D_{\rm evidence},D_{\rm address}$ & Exact-history discrepancy indicator and maximum reconstructed-address error.\\
$\epsilon,\gamma_q$ & Uniform candidate-address error bound and semantic Top-$K$ boundary margin.\\
\bottomrule

\end{longtable}
\endgroup

\FloatBarrier
\Needspace{8\baselineskip}
\section{Host Checkpoints}
\label{sec:hostgrid}
The table lists the reference hosts used in the main comparisons and the checkpoints for Qwen quality scaling and the heterogeneous scale-up chain. Appendix~\ref{sec:implementation_details} describes the implemented reference interfaces.

\begin{table}[!htbp]
\centering\handtablefont
\begin{tabularx}{\linewidth}{@{}L{.23\linewidth}YY@{}}
\toprule
\textbf{Role} & \textbf{Checkpoint family / model card} & \textbf{Planned comparison}\\
\midrule
Qwen quality anchor & Qwen3-4B-Instruct-2507 & Main quality/storage host and the 4B starting point for same-host quality scaling.\\
Qwen dense capacity & Qwen3-4B-Instruct-2507; Qwen3-8B, Qwen3-14B, Qwen3-32B \citep{qwen3models} & Independent same-host quality scaling on four benchmarks; LoCoMo-only 4$\to$8$\to$14$\to$32 migration chain.\\
Mistral reference host & Mistral-7B-Instruct-v0.3 & LoCoMo heterogeneous transfer and target-native rebuilding.\\
Phi reference host & Phi-3-mini-4k-instruct (3.8B) & LoCoMo heterogeneous transfer and target-native rebuilding; execution mode follows Section~\ref{sec:transfer}.\\
Ministral scale-up receiver & Ministral-3-14B-Instruct-2512-BF16 \citep{ministral3card} & First receiving stage after Qwen3-4B-Instruct-2507.\\
Gemma scale-up receiver & gemma-4-26B-A4B-it \citep{gemma4card} & Second receiving stage after Ministral. MoE with 25.2B total and 3.8B active parameters.\\
\bottomrule
\end{tabularx}
\caption{Checkpoint identities for the reference comparisons and Qwen scaling and cross-family transfer. All conditions use text-only inputs.}
\end{table}

\FloatBarrier
\Needspace{8\baselineskip}
\section{Lifecycle and Reference Details}
\label{sec:reference_details}
The tables below expand the revision example, benchmark measurements, reference configuration, and operational contracts. The lifecycle algorithm connects observation writes, evidence reads, snapshots, and model handoff.

\subsection{Revision trace}
\label{sec:trace_details}
Table~\ref{tab:example} expands the versioned-budget example introduced in Section~\ref{sec:overview}.
\begin{table}[!htbp]
\centering\handtablefont
\begin{tabularx}{\linewidth}{@{}L{.13\linewidth}L{.17\linewidth}YL{.25\linewidth}@{}}
\toprule
\textbf{Stage} & \textbf{Host event} & \textbf{Retained evidence} & \textbf{Consequence for later access}\\
\midrule
1 & A observes & Budget 17; explicitly version 1; user/source identity. & Bind the exact value and provenance.\\
2 & Handoff to B; B observes & Budget 18; version 2. Version 1 remains. & Shared addresses may relate the statements; content is not merged.\\
3 & Handoff to C; C observes & Budget 19; version 3. Both earlier records remain. & C adds a record in the same schema.\\
4 & Return to A; query & Latest and version-1 evidence remain available. & The question determines which version must support the answer.\\
\bottomrule
\end{tabularx}
\caption{Running revision trace, adapted from the implementation's versioned-budget observations. Logical answer requirements are not accuracy measurements.}
\label{tab:example}
\end{table}

\FloatBarrier

\Needspace{8\baselineskip}
\subsection{Benchmark measurements}
\label{sec:evaluation_details}
Table~\ref{tab:benchmarks} specifies the tasks and scoring used in Section~\ref{sec:setup}.
\begin{table}[!htbp]
\centering\handtablefont
\begin{tabularx}{\linewidth}{@{}L{.19\linewidth}YY@{}}
\toprule
\textbf{Benchmark} & \textbf{Main measurement} & \textbf{Complementary stress}\\
\midrule
LoCoMo \citep{locomo} & One all-question composite over 1,986 questions, defined in Equation~\ref{eq:locomocomposite}. & Speaker attribution, temporal reasoning, multi-hop recall, abstention.\\
LongMemEval-S \citep{longmemeval} & Auxiliary-judge answer accuracy over 500 questions. & Cross-session memory, temporal questions, updates, missing information.\\
MemOps \citep{memops} & Longitudinal answer accuracy over 2,006 questions; operation/state diagnostics separately. & Memory lifecycle operations and evidence use across an evolving trajectory.\\
MemoryAgentBench \citep{memoryagentbench} & Macro-average across four memory competencies. & Incremental interaction, retrieval, test-time learning, long-range understanding, and handling revised knowledge.\\
BEAM-1M \citep{beam} & Native-category aggregation over 700 questions from 35 histories, including event ordering. & Long-distance evidence and temporal consistency in million-token histories.\\
\bottomrule
\end{tabularx}
\caption{Task coverage and reported metrics. LongMemEval-S and MemOps use the project's auxiliary judges. BEAM uses its category-specific aggregation, with auxiliary judging where required. All scores in the main comparison are on a 0--100 scale.}
\label{tab:benchmarks}
\end{table}

\FloatBarrier
\paragraph{Comparison settings.}
All completed main-quality cells in Section~\ref{sec:quality} use Qwen3-4B-Instruct-2507 on the stated question sets. Each method retains its native read path and computation budget, which can differ across methods. MIGRO uses its fixed task-specific configuration, including the 2000-source-token LoCoMo reader. Target-adapted heterogeneous rebuilding settings are detailed in Appendix~\ref{sec:target_baseline_settings}. Storage replicas are evaluated independently: every quality--storage point pairs the score and bytes of the actual measured state. Same-evidence native-text and no-memory answers are diagnostic controls rather than substitutes for the main methods.

Mistral-7B-Instruct-v0.3 and Phi-3-mini-4k-instruct are the initial heterogeneous targets. MIGRO's reference profile uses 16 relevance seeds and a 2000-source-token evidence cap. Comparisons identify generation allowances, prompts and calibration budgets. Interface calibration uses public examples rather than benchmark answers. Zero-calibration and fixed-example conditions are separated.

\paragraph{Cohorts and uncertainty.}
Questions within a conversation, identity or source history share evidence, so uncertainty is assessed by resampling those clusters. Paired storage comparisons retain the same source clusters, and adjacent and longitudinal MemOps views belong to the same identity cluster. Calibration-seed variation is assessed separately where calibration is compared. Result records identify dataset versions, observation streams, scorers and checkpoints.

The LoCoMo composite equally weights all 1,986 questions and was selected after the runs; Equation~\ref{eq:locomocomposite} defines it. BEAM's 700-question cohort contains 100 previously inspected questions, so the remaining 600-question score is also reported. The BEAM aggregate includes the native event-ordering component. Its 35-history QA cohort is separate from the five-history storage study.

\paragraph{MemoryAgentBench.}
The added benchmark uses Qwen3-4B-Instruct-2507 and reports the unweighted mean of its four competency scores. MIGRO scores 35.16, TransMem scores 22.80, and LSTMem scores 6.56. This aggregate is separate from the LoCoMo all-question score and the task-specific metrics of the other benchmarks.

\paragraph{Aggregate uncertainty and held-out subsets.}
MIGRO's MemOps score covers 100 identities and 2,006 questions, with an identity-cluster 95\% interval of $[64.51,68.23]$. Its 1,772-question unseen-identity subset scores 66.65. BEAM-1M covers 35 histories and 700 questions, with a history-cluster interval of $[54.06,57.90]$. The 600 previously uninspected questions score 55.58, with an interval of $[53.57,57.70]$. The operation and category breakdowns in Section~\ref{sec:failures} identify the remaining difficulty within these aggregates.

\paragraph{Same-host scaling settings.}
The same-host scaling study starts from Qwen3-4B-Instruct-2507 and extends to Qwen3-8B, Qwen3-14B and Qwen3-32B \citep{qwen3models}. The sweep uses fixed non-thinking chat settings, evidence and answer budgets, precision and scoring.

\paragraph{Independent construction.}
For each method and checkpoint, the quality sweep begins with an empty memory, ingests the benchmark history, and answers with that same host. Observation order, access budgets and the public-example calibration allowance are fixed within each family. Each row therefore measures end-to-end quality with a memory prepared for its own host. Table~\ref{tab:hostscaling} and Figure~\ref{fig:scaling}(a) report the available LoCoMo scaling results. LongMemEval-S, MemOps and BEAM-1M remain planned extensions of this same-host sweep. The continuous Qwen migration study is evaluated separately on LoCoMo in Section~\ref{sec:transfer}.

\paragraph{Migration comparisons and diagnostics.}
MIGRO, PlugMem, and Mem0 use State Transfer; TiMem, A-MEM, LSTMem, and Doc-to-LoRA use Full-History Rebuilding. The within-Qwen study uses LoCoMo only and follows a continuous chain from the Qwen training/source host through consecutive receiving stages. Target-native rebuilding results are reused where applicable, while portable methods' same-host and transferred-state scores remain distinct.

\begin{figure}[!htbp]
\centering
\includegraphics[width=\linewidth]{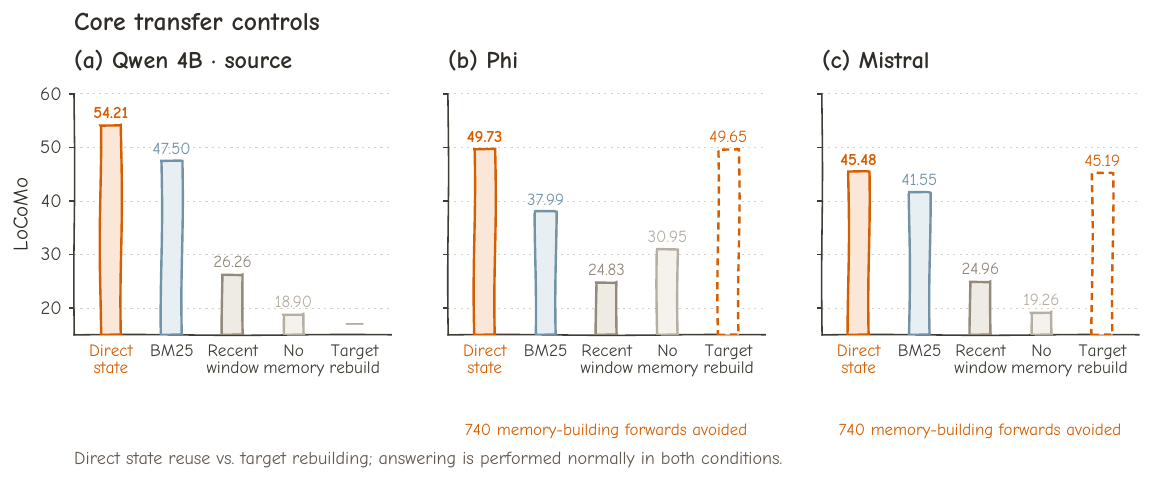}
\caption{Core LoCoMo transfer controls, with 1,986 questions in each of 14 conditions. Host-specific panels compare direct state reuse, BM25 simple memory, recent history, no memory, and target-side rebuilding; the last is inapplicable to the Qwen source reference. Axes span 15--60. Direct reuse exceeds BM25 by the reported paired estimates of 11.74 points on Phi and 3.93 on Mistral, with paired intervals above zero. Reported direct-minus-rebuilding estimates are +0.07 and +0.28, with intervals crossing zero; plotted scores are displayed to two decimals. Each target avoids 740 memory-building forwards through direct reuse; answer generation is performed normally.}
\label{fig:coretransfer}
\end{figure}

For State Transfer methods, additional loss is defined against a matched same-method target-native reference:
\begin{equation}
\Delta_s=Q_{h_s}^{\rm self}-Q_{w_{1:s}\to h_s}^{\rm transferred}.
\label{eq:transferloss}
\end{equation}
Here $Q_{h_s}^{\rm self}$ is the quality of that same method's target-native state under matched history and access settings. This diagnostic separates inherited-state effects from target-model capability. Cross-method receiving-host scores do not measure this loss. Local target-rebuilding controls are available for Phi and Mistral (Figure~\ref{fig:coretransfer}). MIGRO's completed continued-write comparison appears in Figure~\ref{fig:continuedwrite}.

Continued-write tests add observations after handoff and separate earlier-history QA, new-history QA, and cross-boundary integration. The transferred-state-only condition freezes the inherited memory at the boundary; the continued-writing condition appends subsequent observations to that memory; full-history rebuilding constructs target-side memory from both portions of history. The single additional migration study is a LoCoMo heterogeneous scale-up chain: Qwen3-4B-Instruct-2507 $\to$ Ministral-3-14B-Instruct-2512-BF16 $\to$ gemma-4-26B-A4B-it. The two receivers inherit state consecutively. Matched target-native references separate changes in host capability from the effects of state inheritance. State Transfer methods continue from their carried states; rebuilding methods reconstruct from the full history available at each boundary.

\paragraph{Matched-state LoCoMo scoring.}
The matched-state comparisons use the same ten LoCoMo conversations. Each operating point pairs the quality of its evaluated states with their total native bytes. The all-question composite equally weights 1,986 questions:
\begin{equation}
Q_{\rm all}=\frac{1540\,F_{1,\,1:4}+446\,A_5}{1986},
\label{eq:locomocomposite}
\end{equation}
Here $F_{1,\,1:4}$ is token F1 over categories 1--4 and $A_5$ is the category-5 score. The composite was chosen after the runs and is used consistently in the main quality and storage tables. Storage replicas retain their own scores. Conversation-bootstrap intervals keep each conversation's questions together.

\paragraph{Additional operation diagnostics.}
The MemOps operation-detection F1 is 53.42 and provenance support is 63.77. These diagnostics complement the answer-quality breakdown in Table~\ref{tab:failures}.

\begin{table}[!htbp]
\centering\handtablefont
\setlength{\tabcolsep}{5pt}
\begin{tabularx}{\linewidth}{@{}L{.15\linewidth}L{.33\linewidth}>{\centering\arraybackslash}L{.12\linewidth}>{\centering\arraybackslash}L{.10\linewidth}>{\centering\arraybackslash}X@{}}
\toprule
\textbf{Benchmark} & \textbf{Operation / capability} & \textbf{Questions} & \textbf{Score} & \textbf{95\% cluster CI}\\
\midrule
MemOps & Remember & 408 & 73.53 & [70.10, 76.96]\\
MemOps & Forget & 400 & 72.75 & [68.50, 76.75]\\
MemOps & Reflect & 480 & 64.17 & [60.41, 67.92]\\
MemOps & Update & 308 & 62.99 & [57.79, 68.18]\\
MemOps & TrajectoryOps & 410 & 57.80 & [53.17, 62.20]\\
\midrule
BEAM-1M & Abstention & 70 & 55.36 & [46.41, 64.64]\\
BEAM-1M & Contradiction resolution & 70 & 39.64 & [35.71, 43.57]\\
BEAM-1M & Event ordering & 70 & 37.44 & [32.54, 42.71]\\
BEAM-1M & Information extraction & 70 & 66.92 & [57.46, 76.41]\\
BEAM-1M & Instruction following & 70 & 75.71 & [65.83, 84.05]\\
BEAM-1M & Knowledge update & 70 & 65.00 & [55.00, 75.00]\\
BEAM-1M & Multi session reasoning & 70 & 54.39 & [46.85, 62.28]\\
BEAM-1M & Preference following & 70 & 79.64 & [72.14, 86.43]\\
BEAM-1M & Summarization & 70 & 36.61 & [31.22, 41.83]\\
BEAM-1M & Temporal reasoning & 70 & 47.62 & [37.38, 58.57]\\
\bottomrule
\end{tabularx}
\caption{MIGRO operation and capability scores under the Qwen reference configuration. Intervals cluster by MemOps identity or BEAM history. Revision, trajectory reasoning and multi-record synthesis remain more difficult than individual recall or following tasks.}
\label{tab:failures}
\end{table}

\FloatBarrier

\paragraph{Boundary-composition ablation.}
The matched local-interface LoCoMo comparisons are Qwen 4B: 54.21 with the full interface and 53.41 without boundary composition (a 0.80-point decrease); Phi: 49.73 and 48.05 (1.68 points); and Mistral: 45.48 and 39.98 (5.50 points). Mistral retains calibration. These complete-interface scores serve as references for this local mechanism ablation and are separate from the main migration runs in Table~\ref{tab:transfer}. The decreases measure the contribution of boundary composition rather than target-relative transfer loss.

\paragraph{Core transfer controls.}
Fourteen conditions each contain all 1,986 LoCoMo questions. In Qwen 4B source / Phi / Mistral order, direct state reuse scores 54.21 / 49.73 / 45.48; BM25 scores 47.50 / 37.99 / 41.55; recent history scores 26.26 / 24.83 / 24.96; and no memory scores 18.90 / 30.95 / 19.26. Target-side rebuilding scores 49.65 on Phi and 45.19 on Mistral. The reported paired direct-minus-BM25 differences are +11.74 and +3.93, with intervals wholly above zero. The reported paired direct-minus-rebuilding estimates are +0.07 and +0.28, with intervals crossing zero. These supplied paired estimates are retained separately from the scores displayed to two decimal places. Direct reuse avoids 740 memory-building model forwards for each receiver; this count excludes the ordinary answer-generation stage. These local controls share full-mechanism reference scores with the boundary-composition ablation; they do not replace the main migration-chain scores in Table~\ref{tab:transfer}.

\paragraph{Continued-write comparison.}
Figure~\ref{fig:continuedwrite} uses three conditions evaluated under the same receiving host. In old-history / new-history / cross-boundary / overall order, transferred state without additional writes scores 53.52 / 6.66 / 29.28 / 36.94; transfer followed by continued writing scores 51.85 / 53.09 / 35.71 / 45.13; and full-history rebuilding scores 51.55 / 52.90 / 36.61 / 45.19. Continued writing avoids reconstructing the pre-handoff history at the receiver. The supplied paired cross-boundary improvement is 6.44 points with an interval above zero; paired estimates and displayed means can differ slightly through rounding. Under the same setting, PlugMem scores 33.46 with transferred state alone, 37.38 with continued writing, and 36.95 with target-side full-history rebuilding. The displayed overall gain from continued writing is 8.19 points for MIGRO and 3.92 for PlugMem; the cross-method gap increases from 3.48 to 7.75 points.

\paragraph{Mem0 continued-write comparison.}
Each of the three conditions covers all 1,986 LoCoMo questions. Mem0 scores 35.59 with transferred state alone, 43.89 after continued writing, and 44.37 with target-side full-history rebuilding. Continued writing raises the overall score by 8.30 points, leaving a 0.48-point gap to rebuilding.

\paragraph{LSTMem continuation from rebuilt memory.}
LSTMem rebuilds memory from the first half of the history before the continuation comparison. In old-history / new-history / cross-boundary / overall order, rebuilding the first half and continuing to write scores 10.41 / 12.42 / 12.51 / 11.15; retaining only the rebuilt first-half memory scores 10.64 / 12.53 / 12.67 / 11.34; and full-history rebuilding scores 10.16 / 12.71 / 12.52 / 11.16. These are rebuilding-start conditions, in contrast to the transferred-state starts for MIGRO, PlugMem, and Mem0 in Figure~\ref{fig:continuedwrite}. LSTMem's scores remain nearly flat across the three conditions.

\paragraph{Doc-to-LoRA and TransMem continuation.}
In continued-writing / first-half-only / full-history-rebuilding order, Doc-to-LoRA scores 10.98 / 11.04 / 11.32, and TransMem scores 7.04 / 9.60 / 7.04. Additional writing does not increase the displayed overall score for either method in this comparison.

\paragraph{A-MEM and TiMem continuation.}
In continued-writing / first-half-only / full-history-rebuilding order, A-MEM scores 34.01 / 29.67 / 37.87, and TiMem scores 37.17 / 32.30 / 40.92. Continued writing improves their overall scores by 4.34 and 4.87 points, respectively, leaving gaps of 3.86 and 3.75 points to full-history rebuilding. Overall continuation scores are now available for all seven baselines in Figure~\ref{fig:continuedwrite}b.

\paragraph{Incomplete comparisons.}
Matched rebuilding references for the heterogeneous scale-up chain are pending. The outstanding transfer comparisons include broader matched-path diagnostics and additional interface ablations. Category-level continued-writing breakdowns are available for all eight methods in the retained result files. The saved-state pairs for continuation and full-history rebuilding are complete and support Figure~\ref{fig:frontier}. The transfer radar uses common receiving-quality, continuation, and execution-mode dimensions; category-specific analyses remain separate.

\Needspace{8\baselineskip}
\subsection{Reference implementation profile}
\label{sec:implementation_details}
The reference implementation supports Qwen3-4B-Instruct-2507 (A), Phi-3-mini-4k-instruct (B) and Mistral-7B-Instruct-v0.3 (C). Application backbones, the semantic encoder, source embeddings, base lexical mappings, cross-encoder and causal predictor are frozen. Per-user writing changes the retained state. Optional rank-16 residuals are calibrated separately and shared as host-interface assets.

LoCoMo uses bound/support selection for A and C. B uses support selection without discourse-edge admission and its boundary interface. C uses a fixed 64-example calibrated interface. LongMemEval uses a role-coverage reader for A, support without edges for B, and a bound reader for C; the C path has no calibration checkpoint. All use 16 relevance seeds and a 2000-source-token cap. Generation allowances are task-specific: 50 tokens for LoCoMo, 256 for LongMemEval and 384 for the general profile. The experiments identify these task-specific operating points so that changes in access are visible alongside host changes.
\begin{table}[!htbp]
\centering\handtablefont
\begin{tabularx}{\linewidth}{@{}L{.29\linewidth}L{.28\linewidth}Y@{}}
\toprule
\textbf{Component} & \textbf{Reference specification} & \textbf{Scope / consequence}\\
\midrule
Semantic address & $d_a=384$; innovation rank 64; reuse threshold 0.48. & Shared, host-independent organization.\\
Saved address precision & Roots: 4 bits/coordinate, FP16 scale per 64-coordinate group. Innovations: int8 with FP16 scale. & Saved precision is preserved by archival coding.\\
Local writer & At most 4 preceding chronological records, filtered to the same session. Zero-based layer $\max(0,\lfloor2L_h/3\rfloor-1)$. & One current-host local forward per batch; inputs must fit context.\\
Candidate/reranker & Up to 64 candidates; frozen MiniLM cross-encoder, 384-token pair cap, batch 32. & Reference evidence-reader path; candidate policy may vary by declared reader.\\
Seed/source budgets & $K=16$; $T=2000$ source tokens including separators. & Seeds differ from delivered record count; target-context fitting follows.\\
Complementarity & Redundancy coefficient 0.35. & Greedy relevance/diversity selection, not global optimality.\\
Composed transport & 8 source-piece hypotheses; softmax scale 100; word beam 4; position mass threshold 0.5. & Lexical/subword construction; output length is variable.\\
Host residual & Rank 16; frozen backbone. & Interface assets, not per-user observations.\\
Causal archive & Qwen3-4B predictor; BF16, SDPA; 1024-token pages; 24-bit integer CDF; configured batch 128. & Model/dtype/backend/batch schedule enter the decoding contract.\\
Exact format selection & Eligible legacy archive and independently restored causal candidate; charge new public dependency bytes. & Explicit compaction; no latency term in this selector.\\
\bottomrule
\end{tabularx}
\caption{Implementation specifications, not empirical results or architecture-wide optimal settings.}
\label{tab:spec}
\end{table}

\FloatBarrier
\Needspace{8\baselineskip}
\paragraph{Target-built baseline settings.}
\label{sec:target_baseline_settings}
The heterogeneous Full-History Rebuilding comparisons use target-adapted implementations. LSTMem and Doc-to-LoRA fit target-compatible memory interfaces on 1,024 external human HotpotQA examples for four epochs (1,024 updates), with frozen backbones. LoCoMo answers are excluded from this fitting. Doc-to-LoRA retains its released down-projection target set. For the LoCoMo quality and migration comparisons, LSTMem writes every observation and answers from persistent C/D states alone. The 4B reference uses the released adapter, and Qwen 8B, 14B and 32B use target-specific refitting with the same 1,024-example, four-epoch budget. The separate matched-state storage evaluation uses LSTMem's retained-history reader and counts the retained history in its storage total. Phi TiMem uses a context-bounded hierarchy with intermediate summaries capped at 384 tokens and hierarchy inputs capped at 1,536 tokens, while retaining the ordinary 1,024-token final-summary allowance where available. These settings identify the reported target-built variants.

\clearpage
\subsection{Complete memory lifecycle}
\label{sec:lifecycle_algorithm}
Algorithm~\ref{alg:lifecycle} specifies writing, reading, snapshotting, restoration and host replacement as one lifecycle. Accepted writes invalidate reader workspace and cached snapshots. Handoff uses a current snapshot, restoring it through compatible shared dependencies and constructing the receiving access runtime. Ordinary saving supplies a snapshot when the cached one is stale; switching models does not itself request compaction.
\begin{algorithm}[H]
\caption{MIGRO Memory Lifecycle}
\label{alg:lifecycle}
\handtablefont
\textbf{Inputs:} shared dependencies $\mathcal B$; supported interfaces and profiles $\{\Pi_h,\mathcal I_h\}$; chronological observe/query/snapshot/handoff events.\\
\textbf{State:} retained $\mathcal M$, current host $h$, optional cached snapshot $S$, temporary reader workspace $W$.\par\vspace{5pt}
\begin{tabularx}{\linewidth}{@{}r@{\quad}Y@{}}
\toprule
1 & Initialize $\mathcal M$ empty or restore a supplied snapshot; initialize runtime for $h$.\\
2 & \textbf{for each} event $e$ \textbf{do}\\
3 & \quad\textbf{if} $e=\operatorname{Observe}(o)$ \textbf{then}\\
4 & \qquad Form $v=\tau_s(\sigma(o))$; encode the common semantic address.\\
5 & \qquad Select at most the configured chronological past window, retaining same-session records.\\
6 & \qquad Require the local input to fit the host context; forward current host once over past + $o$.\\
7 & \qquad Pool local spans; infer predecessor $\rho$ if the allowed window is nonempty.\\
8 & \qquad Assign/reuse anchor and innovation; append $v$, provenance, writer identity, and $\rho$.\\
9 & \qquad Retain earlier records and bookkeeping; invalidate $S$ and $W$.\\[2pt]
10 & \quad\textbf{else if} $e=\operatorname{Query}(q)$ \textbf{then}\\
11 & \qquad Build/reuse temporary address and lexical workspace from $\mathcal M$.\\
12 & \qquad Generate candidates; rerank; select complementary seeds and eligible support under $K,T$.\\
13 & \qquad Recover exact source IDs and separators; embed with $E_s$; apply $\mathcal I_h$.\\
14 & \qquad While target input + generation reserve exceeds context, remove a delivered unit and rebuild.\\
15 & \qquad Generate answer with frozen host; return answer and evidence metadata.\\
16 & \qquad Clear query-local transport cache; leave $\mathcal M$ unchanged.\\[2pt]
17 & \quad\textbf{else if} $e=\operatorname{Snapshot}(\mathrm{compact},\mathrm{coder})$ \textbf{then}\\
18 & \qquad If not compact, save the cached/default snapshot and continue to the next event.\\
19 & \qquad Otherwise obtain eligible exact legacy candidates, including a valid retained snapshot.\\
20 & \qquad \textbf{if} coder supplied and $\mathcal M$ nonempty \textbf{then}\\
21 & \qquad\quad Encode bounded causal pages, geometry, framing, and both exception layers.\\
22 & \qquad\quad Independently restore new causal candidate; require exact saved-state comparison.\\
23 & \qquad\quad Admit it to the eligible candidates.\\
24 & \qquad \textbf{end if}\\
25 & \qquad Select by actual full bytes + newly required public dependency cost; cache/save $S$.\\[2pt]
26 & \quad\textbf{else if} $e=\operatorname{Handoff}(h')$ \textbf{then}\\
27 & \qquad Get cached $S$, or create the ordinary legacy snapshot if invalidated.\\
28 & \qquad Restore $\mathcal M$ from $S$ with compatible $\mathcal B$; set $h\gets h'$.\\
29 & \qquad Recreate the receiving runtime and workspace; do not replay all history through $h'$.\\
30 & \quad\textbf{end if}\\
31 & \textbf{end for} \hfill\emph{Further Observe events use the receiving host.}\\
\bottomrule
\end{tabularx}
\vspace{4pt}
\parbox{\linewidth}{\footnotesize The algorithm abstracts record-at-a-time events; the implementation can batch local writer forwards. Ordinary save reuses a valid snapshot or selects the default legacy format; only explicit compaction considers the wider candidate set. New causal candidates undergo independent restore-and-compare. Snapshot restoration/index construction may scan the full state. Optional task-specific answer policies are part of $\Pi_h$, not additional writes.}
\end{algorithm}

\FloatBarrier

\Needspace{8\baselineskip}
\subsection{Operational Contracts and Restoration Semantics}
\label{sec:operational_contracts}
Table~\ref{tab:contracts} separates saved-state preservation, access rules, and empirical objectives. Under compatible shared decoding dependencies, the restoration identity is
\begin{equation}
\operatorname{Restore}\!\left(\operatorname{Archive}(\mathcal M_n;\mathcal B);\mathcal B\right)
=\mathcal M_n.
\label{eq:theoryroundtrip}
\end{equation}
Equality is defined over saved source IDs, quantized coordinates, provenance, relations, and historical records. It preserves saved precision rather than recovering pre-quantization vectors.

\begin{table}[!htbp]
\centering\handtablefont
\begin{tabularx}{\linewidth}{@{}L{.21\linewidth}YL{.25\linewidth}@{}}
\toprule
\textbf{Property} & \textbf{Contract and boundary} & \textbf{Status}\\
\midrule
Saved-state recovery & Restore exact source IDs, saved coordinates, provenance and history under compatible decoding dependencies. & Conditional format/coding property; verify decoded candidates.\\
Historical retention & Appending an accepted observation retains earlier logical fields; archive bytes may be repacked. & State/update contract; not crash-safe transactional durability.\\
Read non-mutation & Answering changes temporary caches, not the retained experience. & Runtime contract.\\
Bounded host access & Budgeted evidence for queries; bounded local past for writes. Full restoration/index construction may still scan history. & Access contract; not constant-time migration.\\
Cross-host usefulness & Receiving hosts answer accurately and can use revisions and provenance. & Empirical objective, not implied by exact recovery.\\
Storage advantage & Lower complete bytes at a useful quality/transfer operating point. & Measured quality--storage evidence in Section~\ref{sec:storage}; broader transfer comparisons pending.\\
\bottomrule
\end{tabularx}
\caption{System contracts separate exact preservation, structural access rules, and claims requiring experiments.}
\label{tab:contracts}
\end{table}

\FloatBarrier

Restoration and index construction can traverse the complete state. The archive predictor also processes content during encoding and decoding. These costs enter restoration and deployment accounting separately from rebuilding history through the receiving application's backbone. Public decoding assets remain required dependencies outside the private user payload.

\subsection{Storage Accounting and Shared Dependencies}
\label{sec:storage_accounting}
Persistent bytes, shared method assets and temporary workspace occupy different resource boundaries. Agent-native memory studies motivate this separation by examining representation, extraction, access and maintenance together \citep{agentnative2026}.

Let $L$ count original source-body tokens, excluding metadata added by canonicalization. Complete private bytes and the marginal rate are
\begin{equation}
\begin{aligned}
B_{\rm user}(L)&=B_{\rm address}+B_{\rm content}+B_{\rm provenance}\\
&\quad+B_{\rm index}+B_{\rm framing}+B_{\rm exceptions},\\
\beta(L,\Delta L)&=\frac{B_{\rm user}(L+\Delta L)-B_{\rm user}(L)}{\Delta L}.
\end{aligned}
\label{eq:bytes}
\end{equation}
All retained metadata enters the numerator even though it is outside the source-body denominator. Cold-archive size uses serialized file lengths. A content-only rate substitutes $B_{\rm content}$ and is reported as such. Resident serving state is measured separately.

Equation~\ref{eq:deployment} defines total persistent deployment cost across independent users.
The shared term includes required encoders, tokenizers, lexical assets, rankers, host interfaces and predictor assets. A dependency is charged once when shared, including any newly required predictor. Decoded records, lexical indexes, reconstructed addresses, transport tensors and inference caches contribute temporary workspace.

For $N$ independent memories with mean private size $\overline B_{\rm user}$,
\begin{equation}
\begin{aligned}
B_{\rm deploy}(N)&=B_{\rm shared}+N\overline B_{\rm user},\\
C_{\rm memory}(N)&=\frac{B_{\rm shared}}{N}+\overline B_{\rm user}.
\end{aligned}
\label{eq:amortization}
\end{equation}
The equation exposes two scaling effects. Extending one history incurs its marginal private cost $\beta$. Adding independent memories spreads fixed assets over a larger $N$ while adding one private state per memory. Section~\ref{sec:storage} combines measured state sizes with the required method-asset ledger. The reference causal predictor uses the same Qwen3-4B-Instruct-2507 checkpoint as the common backbone. Its weights and the software runtime are outside this method-specific ledger; a separately introduced predictor would instead enter the shared term once.

Online appends and cold compaction also have different costs. A write invalidates the cached snapshot, while explicit compaction may scan and recode the full state. The current lifecycle provides logical retention and exact saved-state recovery. Crash-consistent transactions, deletion/unlearning and incremental neural recoding are outside the implemented contract.

\paragraph{Persistence-coding ablation.}
The uncompressed archive occupies 2,381,266 B. Existing lossless coding occupies 952,995 B, saving 59.98\% relative to the uncompressed archive. Full encoding selection occupies 869,928 B, saving 63.47\%. These are total archive byte counts for the supplied coding comparison.

\paragraph{Address-table comparison.}
Full float32 addresses occupy 9,036,512 B and score 54.67 on LoCoMo. Anchor + Residual occupies 802,550 B and scores 54.21; Anchor only occupies 560,516 B and scores 51.73. Anchor + Residual reduces address-table size by 91.1\% relative to float32, with a 0.46-point quality gap. The Anchor + Residual and Anchor-only quality measurements are shared with the performance ablation. Address-table sizes and the archive totals above refer to different measured objects and are accounted for separately.

\paragraph{Matched-state comparison.}
MIGRO attains the highest matched-state LoCoMo score, 54.13, with 1.115 MB across ten states (Table~\ref{tab:storage}; Figure~\ref{fig:storagequality}a). Against Mem0, the strongest matched-state quality baseline, it improves the score by 3.68 points while reducing bytes by 99.3\%; the paired 95\% interval for the score difference is $[2.49,5.06]$. TransMem uses 1.200 MB and scores 45.96; MIGRO gains 8.17 points while using 7.1\% less storage. Against LSTMem, MIGRO uses 40.4\% less space and scores 13.50 points higher. The compact representation therefore retains evidence useful to the measured answers.

\paragraph{Measurement scope and history prefixes.}
The storage line tests whether useful answers coexist with a compact state, how that state grows with history, and how fixed assets amortize across independent memories. All sizes use decimal MB ($10^6$ bytes). Native storage includes recoverable user-specific content, indexes, provenance, retained history and learned state. Each method keeps its native format; MIGRO uses its built-in codec, with no external compressor added to the baselines. Fixed method assets enter the shared-resource comparison separately.

LoCoMo prefixes also show the cost of continued writing (Figure~\ref{fig:storagequality}b--d). Each prefix compares the same eligible conversations, whose count falls as the requested history grows. Marginal rates are computed within conversations before averaging. MIGRO's second-half cost is 4.207 bytes/token, compared with 4.215 for TransMem, 4.900 for Doc-to-LoRA, 8.847 for LSTMem, 49.385 for TiMem, 113.668 for A-MEM, 410.676 for PlugMem and 820.634 for Mem0.

Five natural BEAM histories extend the storage comparison to one million source tokens each. The seven complete trajectories have 225 checkpoints per method: 44 common thresholds and a full endpoint for each history. TransMem's full prefix series is complete; Doc-to-LoRA has all five paired core histories for the table, while its optional dense series contains 35 measurements. Actual endpoint lengths average 1,000,100.4 source tokens. Three histories extend development streams and two extend confirmation streams. This cohort measures history growth separately from the 35-history QA study.

The late-history rate averages the within-history bytes added between actual 500K and 1M boundaries:
\begin{equation}
\overline\beta_{\rm late}=\frac{1}{5}\sum_{j=1}^{5}
\frac{B_{{\rm user},j}(L_{j,1\rm M})-B_{{\rm user},j}(L_{j,500\rm K})}
{L_{j,1\rm M}-L_{j,500\rm K}}.
\label{eq:latemarginal}
\end{equation}

\paragraph{Local and half-history marginal rates.}
TransMem's 975K--1M local marginal is 7.240 bytes/token; its matched 500K--1M mean is 7.778 bytes/token. Doc-to-LoRA's private LoCoMo states total 283.14 MB and score 14.96; their second-half marginal cost is 4.9 bytes/token. Its five-history BEAM mean is 56.625 MB at one million tokens. Over 975K--1M, the reported local marginal is 0.000040 bytes/token, whereas the 500K--1M mean is 28.313 bytes/token. A near-zero final-window increment can therefore coexist with a much larger accumulated state. Table~\ref{tab:growth} and the storage radar use the shared 500K--1M comparison window; the final-window value is retained as a separate diagnostic. Intervals for these two methods use 4,000 bootstrap resamples of the five histories. Doc-to-LoRA's endpoint byte count is identical across the five histories, yielding equal lower and upper empirical interval bounds.

\paragraph{Paired continuation states.}
Figure~\ref{fig:frontier} measures the saved states used by the continued-writing and full-history rebuilding runs. The paired cohort contains ten conversations and 1,986 identical question IDs in every arm. Native payloads include MIGRO snapshots; PlugMem's persisted graph and session metadata; Mem0's vector store and history database; A-MEM's notes and embeddings; TiMem's hierarchy and vectors; and LSTMem's saved C/D matrices. Doc-to-LoRA retains its per-chunk factors in this experiment. TransMem's selected view retains its context, overflow tensors and native metadata. Other experimental control views, predictions, build logs and fixed shared weights are excluded. Completion headers use the existing LoCoMo accounting convention. All plotted continuation points use the Mistral receiving-host setting and their corresponding native saved representations.

\paragraph{Shared-deployment calculation.}
Equation~\ref{eq:amortization} combines the measured million-token state means with one shared method-resource bundle. MIGRO's required assets occupy 297.166 MB. The required shared bundles for TransMem and Doc-to-LoRA are 1.720 GB and 1.684 GB, respectively. Across $N=1{,}000$ independent memories, the shared term contributes 0.297 MB per memory:
\begin{equation}
C_{\rm memory}(1{,}000)
=\frac{297.166}{1{,}000}+1.857
=\mathbf{2.154}\ \text{MB/memory}.
\label{eq:sharedresult}
\end{equation}

Figure~\ref{fig:sharedstorage} distinguishes the current deployed inventory, including unused weights, from required method resources. Both exclude the common Qwen backbone and runtime. Bands propagate uncertainty in the five-history state mean through the shared-resource arithmetic.

\subsection{Runtime Complexity and Operating Conditions}
\label{sec:runtime_complexity}
Construction, retrieval and generation can shift resource costs independently of archive size \citep{memoryworkloads2026}. With $a$ anchors, address assignment scans residual errors in $O(a(d_a-r))$ arithmetic. Host-local writing is bounded in record count, with compute still determined by the admitted token lengths. Inputs must fit the host context; an over-context local batch is rejected.

Reader construction can traverse all $n$ records to reconstruct addresses and lexical statistics. The reference dense scan costs $O(nd_a)$, followed by bounded cross-encoder scoring and greedy seed selection. Transport adds source-vocabulary similarity and target tokenization. Generation then depends on the expanded evidence and answer length. A fixed evidence cap bounds delivered content without eliminating these preparation costs.

Causal encoding and decoding process the complete predictor-token stream. Page-local context bounds each predictive window, while total reconstruction work grows with the number of pages. Compact persistence thus trades archive bytes against shared prediction and restoration work. The cold archive and the temporary serving representation serve different roles: one retains recoverable history economically, and the other supports active access.

\Needspace{8\baselineskip}
\subsection{Capability Profiles and Joint Frontiers}
\label{sec:joint_details}
\paragraph{Capability definitions and normalization.}
Figure~\ref{fig:radar} summarizes the three lines in separate compact panels. The performance panel highlights six selected strengths:
\begin{itemize}[leftmargin=*,itemsep=2pt]
\item \textbf{Cross-session reasoning:} the multi-session question type in LongMemEval-S.
\item \textbf{Temporal understanding:} the temporal-reasoning question type in LongMemEval-S.
\item \textbf{Knowledge adaptation:} the knowledge-update question type in LongMemEval-S.
\item \textbf{Learning from examples:} exact match on MemoryAgentBench's CLINC150 task with a 7,050-example demonstration context, using the common format-corrected scorer.
\item \textbf{Long-horizon recall:} information extraction in BEAM-1M.
\item \textbf{Scale-up quality:} mean same-host LoCoMo at Qwen 8B, 14B, and 32B from Table~\ref{tab:hostscaling}.
\end{itemize}
Each performance spoke has its own scale, with an outer limit 5\% above its best displayed score. All methods on a spoke share that limit. All eight methods have complete measurements for these settings. Table~\ref{tab:mainquality} provides the full benchmark comparisons.

Abstention is reported separately and remains part of the full BEAM-1M aggregate in Table~\ref{tab:mainquality}. The available abstention scores are 55.36 for MIGRO, 92.50 for LSTMem, 43.57 for TiMem, 82.86 for PlugMem, 45.71 for Mem0, 91.79 for TransMem, and 75.71 for Doc-to-LoRA. The BEAM category scores accompany the authoring files.

\begin{figure}[!htbp]
\centering
\includegraphics[width=\linewidth]{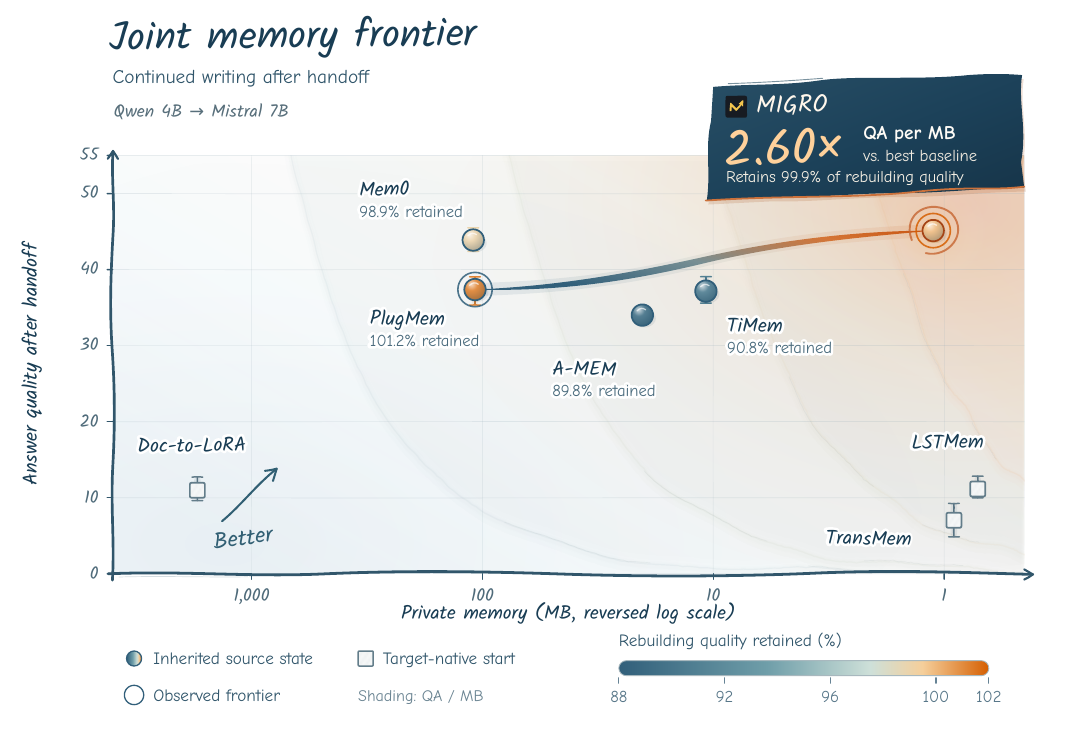}
\caption{Joint frontier on matched memory states after Qwen-to-Mistral handoff and continued writing. Higher answer quality and smaller private memory place a method farther toward the upper right. The ink stroke connects the two observed frontier points. Marker color shows quality retained relative to same-method rebuilding. Squares mark target-native references, shading follows QA/MB, and whiskers show 95\% conversation-bootstrap quality intervals where available.}
\label{fig:frontier}
\end{figure}

The transfer panel highlights six aspects of using memory across hosts:
\begin{itemize}[leftmargin=*,itemsep=2pt]
\item \textbf{Upgrade readiness:} LoCoMo quality at Qwen 8B, 14B, and 32B, divided by the best displayed score at each receiver and then averaged.
\item \textbf{Cross-family reach:} the corresponding average over Mistral 7B and Phi.
\item \textbf{History continuity:} overall LoCoMo quality using only the first-half memory before subsequent writes.
\item \textbf{Continued learning:} overall LoCoMo quality after extending that memory with the second-half observations.
\item \textbf{Portable efficiency:} continued-writing quality per private MB from the matched receiving-host states in Figure~\ref{fig:frontier}.
\item \textbf{Replay-free reuse:} execution mode in Table~\ref{tab:transfer}, coded as 1 for State Transfer and 0 for Full-History Rebuilding.
\end{itemize}
The first two dimensions use the receiving scores under each method's execution mode. TransMem's Qwen entries use the target-native construction and read results in Table~\ref{tab:hostscaling}. The continuation dimensions use Figure~\ref{fig:continuedwrite}. The binary axis records the primary migration setup in Table~\ref{tab:transfer}. Each spoke has a separate outer limit 5\% above its best displayed value. All eight methods have complete profiles.

The storage panel covers the cost of keeping, using, growing and deploying memory:
\begin{itemize}[leftmargin=*,itemsep=2pt]
\item \textbf{Compact state.} Total private MB across the ten LoCoMo conversation states.
\item \textbf{Quality density.} Matched-state LoCoMo score per private MB.
\item \textbf{Lean appends.} Added private bytes per source token over the second half of each LoCoMo conversation.
\item \textbf{Long-horizon compactness.} Private MB per million-token BEAM history.
\item \textbf{Economical growth.} Added private bytes per source token from 500K to 1M tokens in BEAM.
\item \textbf{Deployment economy.} MB per million-token BEAM memory at 1,000 independent memories, including the required shared resources amortized across them.
\end{itemize}
Table~\ref{tab:storage} reports the matched private-state measurements.

The storage panel includes all eight methods with complete six-dimensional profiles. MIGRO has the smallest displayed private footprints, the highest LoCoMo score per private MB, and the lowest measured marginal costs over the matched half-history windows and BEAM deployment cost.

QA yield per MB uses $u_{mj}=z_{mj}/\max_{m'}z_{m'j}$, and a positive storage cost uses $u_{mj}=\min_{m'}C_{m'j}/C_{mj}$. Transfer values are divided by each spoke's best displayed value for profile accounting; the display margin does not enter that normalization. All-zero or undefined references remain unranked. The storage display's outer radius is 1.05, leaving a margin beyond the best normalized efficiency of 1.00. Raw inputs accompany the source package.

The profiles summarize capabilities across tasks. The frontier preserves the measured relationship between quality, transfer and storage for one complete operating point.

\paragraph{Joint quality--portability--storage frontier.}
Figure~\ref{fig:frontier} pairs continued-writing quality under Mistral with the native states used to answer the same 1,986 questions across ten conversations. The setting combines quality, portability and private storage at one operating point; fixed shared resources are accounted for separately.

Color gives rebuilding-relative quality, \(r_m=Q_m^{\mathrm{continue}}/Q_m^{\mathrm{rebuild}}\), against each method's matched full-history rebuilding reference. Among inherited-state methods, the joint frontier maximizes answer quality and $r_m$ while minimizing private bytes. Values above one retain observed gains over rebuilding; soft shading follows answer yield per MB.

MIGRO combines the highest continued-writing quality with the smallest inherited-state footprint, retaining 99.9\% of rebuilding quality. Its answer yield per MB is 2.60 times the best baseline across all eight methods. MIGRO and PlugMem form the observed joint frontier: PlugMem retains 101.2\% of its own rebuilding quality at a larger footprint and lower absolute quality. Useful experience can therefore remain accessible and economical as the receiving host extends it.

MIGRO, PlugMem, Mem0, A-MEM and TiMem inherit Qwen prefix states and append under Mistral. Reference squares denote LSTMem and Doc-to-LoRA's target-built prefixes and TransMem's view rebuilt on append. These continuation operations differ from the primary migration comparisons in Table~\ref{tab:transfer}; Appendix~\ref{sec:storage_accounting} details native-state accounting.

\Needspace{8\baselineskip}
\section{Theoretical and Algorithmic Details}
\label{sec:technical_details}
\subsection{Proofs of the Preservation and Stability Propositions}
\label{sec:proposition_proofs}
\begin{proof}[Proof of Proposition~\ref{prop:irreversibility}]
There is a question $q^\star$ with $\operatorname{Ans}(q^\star,R_1)\ne\operatorname{Ans}(q^\star,R_2)$. Its query and encoded memory are identical in the two cases. A deterministic readout produces the same answer; a randomized readout has the same output distribution. Neither can guarantee both required answers from that same input.
\end{proof}

\Needspace{8\baselineskip}
\begin{proof}[Proof of Proposition~\ref{prop:topkstability}]
Cauchy--Schwarz gives
\begin{equation}
|s_i(q)-\widehat s_i(q)|
=|\langle k_q,k_i-\widehat k_i\rangle|
\le\|k_q\|_2\|k_i-\widehat k_i\|_2\le\epsilon.
\label{eq:scoreperturbation}
\end{equation}
For any original Top-$K$ member $i$ and any nonmember $j$, the perturbed separation is at least
$\widehat s_i-\widehat s_j\ge s_i-s_j-2\epsilon\ge\gamma_q-2\epsilon>0$.
Every member therefore remains above every nonmember.
\end{proof}

\Needspace{8\baselineskip}
\subsection{An Information-Theoretic Illustration of Deferred Selectivity}
\label{sec:binary_example}
Consider a history with $N_f$ independent binary facts and a future question family that can request any individual fact. Every pair of different binary histories differs at some position, and a question about that position distinguishes them. A query-independent, zero-error fixed-length encoding must therefore distinguish all $2^{N_f}$ histories, requiring at least $N_f$ bits \citep{cover2006information,kushilevitz1997communication}.

Once the question supplies the requested position, one bit communicates the corresponding value. Preserving information for a family of possible future questions and delivering information for one question have different information requirements. MIGRO retains a recoverable canonical history and constructs compact query-conditioned evidence packets from it.

\Needspace{8\baselineskip}
\subsection{Host-Local Reference Inference}
\label{sec:reference_inference}
Before addressing an observation, shared canonicalization incorporates available speaker and date information with deterministic temporal annotations. The public encoder then supplies its semantic address. For observation $i$, $\mathcal W_i$ contains the same-session records admitted from a bounded chronological past window. The current host $h_i$ processes these records together with the incoming observation. For each observation span $I_{i,j}$, the pooled normalized vector and predecessor choice are
\begin{equation}
b_{i,j}^{(h_i)}=\operatorname{norm}\!\left(
\frac{1}{|I_{i,j}|}\sum_{t\in I_{i,j}}H_t^{(h_i,\ell_w)}
\right),\qquad
\rho_i=\arg\max_{j\in\mathcal W_i}
\langle b_{i,i}^{(h_i)},b_{i,j}^{(h_i)}\rangle .
\label{eq:reference}
\end{equation}
An empty window gives no predecessor. Every vector is computed by the same current host, including when a past record was written under a different host. Only the resulting observation reference and provenance persist. These local hints supplement independently retained source bindings. A cross-session revision can remain unlinked while its source binding remains intact.

\Needspace{8\baselineskip}
\subsection{Complementary Evidence Selection}
\label{sec:complementary_selection}
Let $\mathcal C_q$ be the reranked candidate set and $S$ the local set of selected relevance seeds. MIGRO uses an MMR-style rule \citep{mmr} to balance query relevance and semantic redundancy:
\begin{equation}
i^\star=\arg\max_{i\in\mathcal C_q\setminus S}
\left[s(q,i)-\lambda\max_{j\in S}
\langle\widehat k_i,\widehat k_j\rangle\right].
\label{eq:mmr}
\end{equation}
The penalty discourages repeated selection from the same semantic neighborhood. Eligible local and reference-linked support is admitted by the reader policy within the delivered-token budget. The seed limit differs from the delivered-record count, and source cues used for scoring are distinguished from exact content sent to the host. The final packet retains source attribution and chronological order.

A delivered unit can be a complete observation or an exact source span under the selected policy. The relevance-seed limit $K$ controls primary selection, while admitted support can increase the delivered-record count within the same source-token budget.

\Needspace{8\baselineskip}
\subsection{Lexical Transport and Host Input Assembly}
\label{sec:lexical_transport}
Table~\ref{tab:spec} specifies the reference dimensions, quantization, local window and budgets. The access profile determines how that state is used for each task and host. In particular, Phi's boundary residual changes an assistant boundary position, while the ordinary memory residual changes transported evidence embeddings. These interfaces share a role but act at different locations.

Selected evidence must also enter the receiving model's computation. MIGRO adapts it at the input boundary while keeping canonical source bindings independent of the target vocabulary. Exact source IDs first give public embeddings $Z_q=E_s(v_{\mathcal E_q})$. Transport interprets source pieces as weighted lexical hypotheses, composes groups across subword boundaries, and represents those hypotheses with the receiving tokenizer and embedding table.

For lexical group $g$, let $\Omega_g$ contain hypotheses $w$ with weights $\alpha_{gw}$. Its transported embedding at position $t$ is
\begin{equation}
G_{g,t}^{(h)}=
\frac{\sum_{w\in\Omega_g}\alpha_{gw}\,
\mathbf 1[t\le|\tau_h(w)|]\,E_h(\tau_h(w)_t)}
{\sum_{w\in\Omega_g}\alpha_{gw}\mathbf 1[t\le|\tau_h(w)|]}.
\label{eq:transport}
\end{equation}
Positions with insufficient hypothesis mass are omitted. The representation can change sequence length and embedding width to match the receiver's tokenizer and input space \citep{kudo2018sentencepiece}. This lexical construction produces continuous target-compatible inputs while the original source bindings remain in persistent memory.

The host assembly operator combines transported evidence with the query and processes it through the frozen layers:
\begin{equation}
H_0^{(h)}=\mathcal I_h(q,Z_q;\eta_h),\qquad
H_{\ell+1}^{(h)}=F_{\theta_h,\ell}(H_\ell^{(h)}).
\label{eq:assembly}
\end{equation}
Optional parameters $\eta_h$ adapt transported memory embeddings or an assistant input boundary, depending on the host. Calibration uses public human-authored examples while the application backbone remains frozen \citep{li2021prefix,hu2022lora}. It belongs to the access interface and is accounted for separately from private experience.

Target tokenization can expand the source packet. The runtime reserves generation space, checks the assembled input against the target context, and removes delivered units and rebuilds the input until it fits. Retrieval and transport thereby connect exact retained evidence to bounded host-specific computation, while indexes, transported embeddings, and generation caches remain temporary.

\Needspace{8\baselineskip}
\subsection{Causal Coding and Exact Archive Selection}
\label{sec:causal_archive}
\label{sec:exact_decoding}
\paragraph{Exact decoding conditions.}
Encoding and decoding use matching predictor and coding-tokenizer versions, integer-CDF convention, page framing, numerical precision, attention backend, and batch schedule. Surface and source-ID exceptions retain tokenizer-sensitive distinctions. Each new causal candidate is independently restored and compared with the saved state before admission.

MIGRO reduces persistent bytes through reversible causal coding. A frozen predictor assigns probabilities within bounded, same-scope pages, and arithmetic coding converts them into a recoverable bitstream \citep{witten1987arithmetic,compression}. Let $y_{p,t}$ be a predictor-vocabulary token, $c_{p,t}=(\mathtt{BOS},y_{p,<t})$ its decoded page-local context, and $\widetilde p_\phi$ the realized integer-CDF probability. With framing $f_p$, encoding and decoding satisfy
\begin{equation}
b_p=\operatorname{ACEnc}_{\widetilde p_\phi}(y_p\mid f_p),\qquad
\operatorname{ACDec}_{\widetilde p_\phi}(b_p\mid f_p)=y_p.
\label{eq:codec_roundtrip}
\end{equation}
Equation~\ref{eq:coding} gives the associated content-length relation.
Predictive context resets at page boundaries. Framing reconstructs fixed separators; arithmetic coding represents the remaining positions. Complete archive bytes include framing and exceptions as well as the coded content.

Two exception layers preserve distinctions across tokenization boundaries. Surface exceptions retain content that the coding tokenizer cannot reproduce exactly. Source-ID exceptions retain the canonical sequence when recovering the surface and retokenizing would change its source IDs. Saved addresses, provenance, reference relations, and historical records accompany the content.

Each new causal snapshot is independently decoded and compared with the saved fields before acceptance. The decoding configuration is specified in Appendix~\ref{sec:exact_decoding}. A validated legacy lossless representation remains eligible. Among eligible exact archives, the system selects
\begin{equation}
c^\star=\arg\min_{c\in\mathfrak C_{\rm exact}}
\left(|\operatorname{Enc}_c(\mathcal M_n)|+B_{\rm newdep}(c)\right).
\label{eq:selection}
\end{equation}
using complete encoded size and newly required shared dependency bytes. This evaluates recoverable storage rather than content likelihood alone. Required predictors and other decoding assets enter the cost when introduced by the archive.

Compaction is explicit. It can reorganize and recode the physical archive while preserving the logical state used by writing and reading. Evidence and saved addresses define memory identity; causal coding determines how economically that state is retained.

\end{document}